\documentclass[acmlarge]{acmart}

\usepackage[capitalise]{cleveref}
\usepackage{graphicx} % Required for inserting images
\usepackage{multirow}
\usepackage{svg}
\usepackage{subcaption}  
\usepackage{pifont}

\usepackage{subcaption}
\AtBeginDocument{%
  }

\setcopyright{acmlicensed}
\copyrightyear{2018}
\acmYear{2018}
\acmDOI{XXXXXXX.XXXXXXX}

\acmISBN{978-1-4503-XXXX-X/18/06}

\acmJournal{IMWUT}
\acmYear{2024} \acmVolume{1} \acmNumber{1} \acmArticle{1} \acmMonth{1} \acmPrice{15.00}\acmDOI{10.1145/3459666}

\begin{document}

\title{ActivityNarrated: An Open-Ended Dense Signal Captioning Method for Wearable Sensor-based Human Activity Understanding}

\author{Lala Shakti Swarup Ray}
\orcid{0000-0002-7133-0205}
\affiliation{%
  \institution{DFKI}
  \city{Kaiserslautern}
  \country{Germany}}
\email{lala_shakti_swarup.ray@dfki.de}

\author{Mengxi Liu}
\affiliation{%
  \institution{DFKI}
  \city{Kaiserslautern}
  \country{Germany}}
\email{mengxi.liu@dfki.de}

\author{Alcina Pinto}
\affiliation{%
  \institution{RPTU and DFKI}
  \city{Kaiserslautern}
  \country{Germany}}
\email{alcina_alfred.pinto@dfki.de}

\author{Deepika gurung}
\affiliation{%
  \institution{RPTU and DFKI}
  \city{Kaiserslautern}
  \country{Germany}}
\email{Deepika.Gurung@dfki.de}

\author{Daniel Geißler}
\affiliation{%
  \institution{DFKI}
  \city{Kaiserslautern}
  \country{Germany}}
\email{daniel.geissler@dfki.de}

\author{Paul Lukowicz}
\affiliation{%
  \institution{RPTU and DFKI}
  \city{Kaiserslautern}
  \country{Germany}}
\email{Paul.Lukowicz@dfki.de}

\author{Bo Zhou}
\email{bo.zhou@dfki.de}
\affiliation{%
  \institution{RPTU and DFKI}
  \city{Kaiserslautern}
  \country{Germany}
}


\renewcommand{\shortauthors}{Ray et al.}

\begin{abstract}
Wearable human activity recognition (HAR) has made steady progress, yet much of this progress remains grounded in fixed-window, closed-set classification benchmarks. 
This formulation is poorly matched to everyday behavior, where activities are open-ended, unscripted, personalized, variable in duration, and often compositional. 
To address this mismatch, we introduce ActivityNarrated, an open-ended narrative paradigm for language-grounded wearable activity understanding. 
We formulate this setting as dense sensor signal captioning with a comprehensive benchmark protocol that measures temporal localization, caption quality, sensor-language alignment, conventional closed-set classification as a downstream diagnostic, and additional robustness measures. 
We further present ActNarrator, a 3-stage architecture that discretizes continuous IMU signals into reusable motion tokens and uses an external frozen small language model to generate open-vocabulary activity captions. 
Experiments show that our method provides high quality dense sensor captioning with superior adaptivity and robustness, enabling various downstream tasks by turning sensor-based human activity understanding into sensor-grounded text-level reasoning.
This includes downstream classification where ActNarrator outperforms state-of-the-art HAR models by 3.8 - 31.6 \% in Macro-F1.
This paradigm also enables novel activity understanding capabilities such as complex question-answering over long time horizons.
\end{abstract}




\received{20 February 2007}
\received[revised]{12 March 2009}
\received[accepted]{5 June 2009}

\begin{teaserfigure}
\centering
  \includegraphics[width=\textwidth]{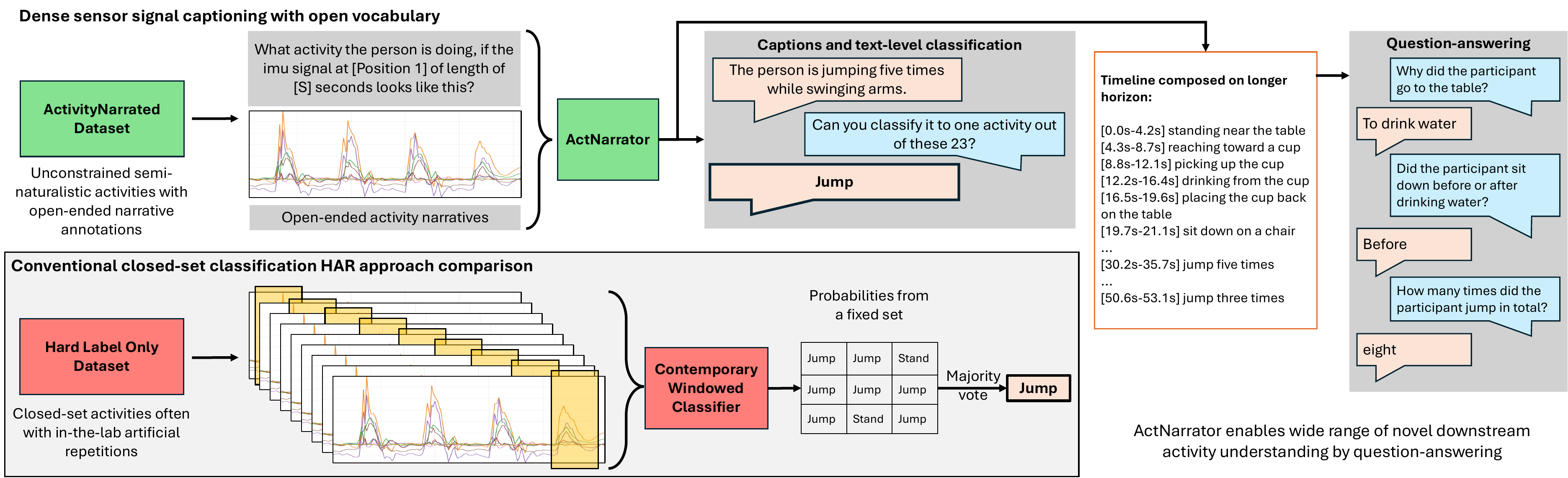}
  \caption{A conceptual shift from windowed, closed-set classification to open-ended activity narrations, where wearable sensor streams are interpreted as semantically rich activity narratives rather than forcefully mapped into fixed labels}
  \label{fig:closed_comp}
\end{teaserfigure}

\maketitle
\section{Introduction}
\label{sec:intro}

Wearable Human Activity Recognition (HAR) has made consistent progress on standard benchmarks \cite{guan2017lstmensemble, abedin2021attend, radu2018multimodal, peng2018aroma}. 
Much of this progress, however, remains tied to a closed-set formulation: participants perform scripted activities, sensor streams are segmented into fixed-size temporal windows, devices are placed at standardized body locations, and models predict one label from a predefined taxonomy \cite{murahari2018attentionISWC, guan2017lstmensemble, tang2021selfhar}. 
This formulation enables controlled comparison, but it only partially reflects everyday activity behavior, where users, sensor placements, activity durations, and semantic descriptions vary substantially.

Everyday behavior is open-ended, personalized, long-tailed, and often compositional. 
A person may walk while gesturing, cook while conversing, or transition fluidly between actions whose boundaries do not align with fixed temporal windows \cite{vaizman2018contextinthewild}. 
The same sensor evidence may also support multiple valid descriptions at different semantic levels, such as ``reaching forward,'' ``picking up a cup,'' or ``preparing coffee.'' 
Expanding the label set or collecting more scripted examples can improve coverage, but it does not remove the closed-world assumption that relevant activities can be enumerated in advance.
Many wearable applications require activity understanding beyond fixed-label prediction. 
Assistive technologies, longitudinal health monitoring, context-aware interfaces, and in-the-wild behavioral studies require systems that can describe what behavior is occurring, when it starts and ends, and how it varies across users, contexts, and sensing configurations \cite{wang2024ubiphysio, vaizman2018contextinthewild}.

We therefore reformulate wearable HAR as open-ended activity narration. 
Instead of classifying fixed windows, we model continuous wearable sensor streams as evidence for temporally localized natural-language descriptions. 
Under this view, wearable activity understanding becomes a dense sensor--language event captioning problem: given one or more continuous IMU streams, a model should identify activity intervals and describe each interval using open-vocabulary language.
Recent work has begun to connect wearable sensing with language models, using language as an interface for querying, summarizing, and transferring activity representations. 
However, many existing approaches still inherit assumptions from closed-set HAR: language is often derived from class names or templates, inputs are pre-segmented into homogeneous windows, evaluation remains tied to known labels, and sensor placement is usually standardized. 
As a result, prior sensor--language approaches do not fully address the combined challenge of open-ended semantics, variable-duration events, heterogeneous sensor placements, and continuous-stream inference.
Addressing this challenge requires rethinking data, evaluation, and model design together. 
Datasets should capture unscripted behavior rather than only repeated executions of predefined activities; evaluation should remain meaningful when multiple descriptions are valid for the same interval; and models should handle continuous streams, heterogeneous body placements, and partially observed sensor configurations. 
These requirements are difficult to satisfy with prevailing wearable HAR datasets and evaluation protocols \cite{tang2021selfhar, abedin2021attend, guan2017lstmensemble}.

We introduce ActivityNarrated, a benchmark and formulation for open-ended, language-grounded wearable activity understanding. 
ActivityNarrated pairs multi-position IMU sensing with time-aligned natural-language descriptions collected during semi-naturalistic activity sessions. 
The dataset contains recordings from 22 participants wearing up to 15 body-worn IMUs, together with expert annotations, participant narrations, auxiliary weak labels, and a compact 23-class taxonomy for closed-set diagnostic evaluation. 
This design supports evaluation under cross-subject, cross-position, and missing-sensor settings while treating closed-set classification as a diagnostic rather than the primary task.
Narrative supervision is not intended as a cheaper replacement for fixed-label annotation. 
Rather, it exposes semantic variation that fixed taxonomies suppress, including differences in granularity, wording, intent, and compositional structure. 
Accordingly, ActivityNarrated is intended as a benchmark-oriented formulation for studying sensor--language activity understanding. 
We also report practical constraints of this protocol, including narration effort, recording awareness, and privacy considerations.

\begin{figure*}[t]
    \centering
    \includegraphics[width=\linewidth]{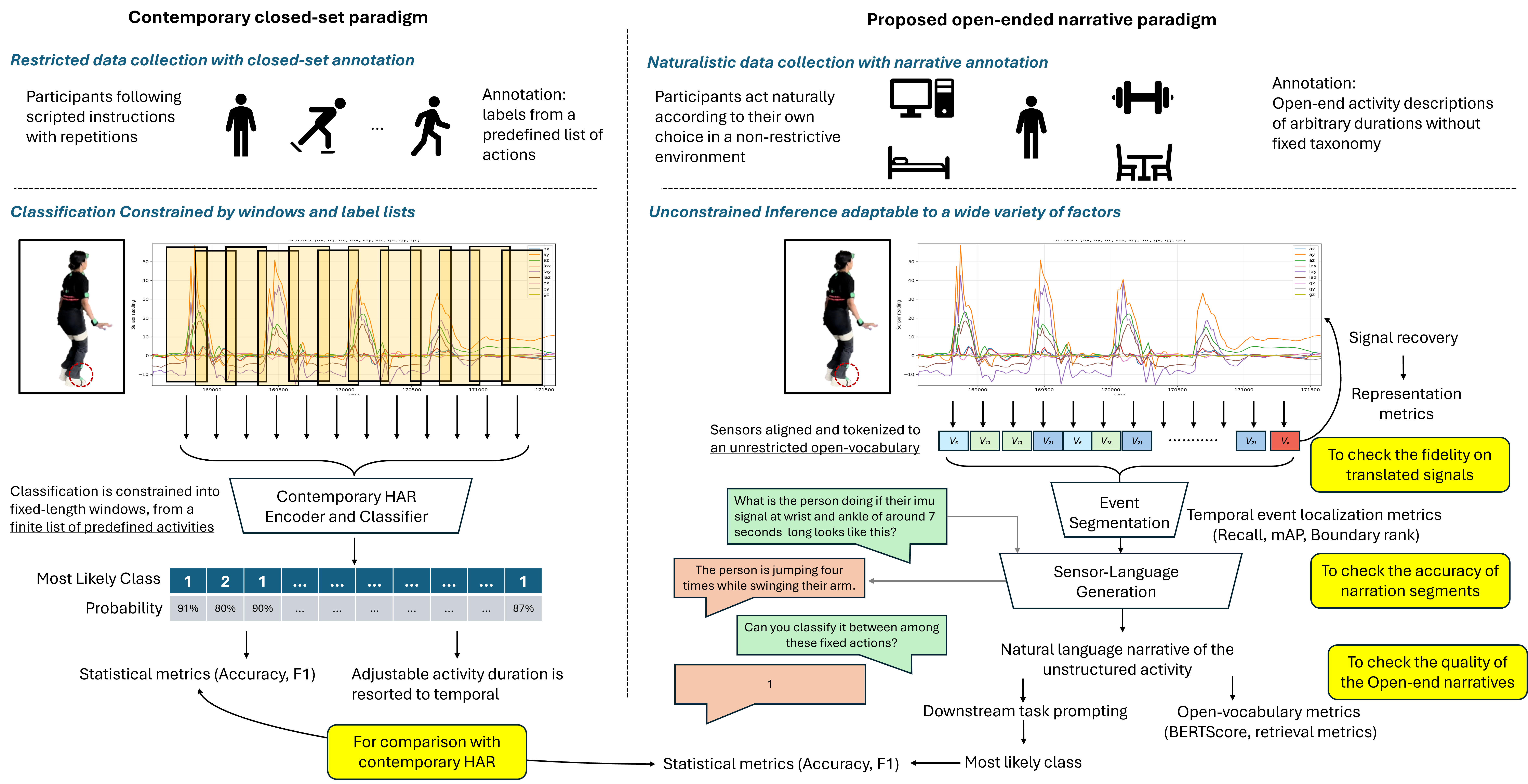}
    \caption{
    Conceptual comparison between prevailing wearable HAR pipelines and the open-vocabulary paradigm introduced in this work. 
    Conventional pipelines typically rely on fixed-size windows, standardized sensor layouts, and closed-set classification, which can limit evaluation under sensor shift, missing sensors, and long-tailed behaviors. 
    In contrast, the proposed paradigm models continuous, multi-position sensor streams through temporally localized natural-language descriptions, allowing closed-set HAR classification to be evaluated as a downstream diagnostic.
    }
    \label{fig:concept_comparison}
\end{figure*}

We further present ActNarrator, a reference architecture for dense sensor--language event captioning from wearable IMU streams. 
ActNarrator discretizes continuous IMU signals into reusable motion tokens, processes long-context multi-position sensor inputs, and predicts activity intervals with open-vocabulary captions. 
Experiments evaluate localization, captioning, retrieval, closed-set diagnostics, held-out activity generalization, timeline question answering, and robustness to subject, placement, and sensor-availability shifts.

This paper makes the following contributions:

\begin{itemize}
    \item \textbf{Open-ended wearable activity benchmark and evaluation protocol}: We reformulate wearable HAR as dense sensor--language event captioning and introduce an evaluation protocol that measures activity event segmentation, open-vocabulary captioning, closed-set diagnostic classification, timeline question answering, and robustness under subject, position, and missing-sensor shifts.
    
%    \item \textbf{ActivityNarrated dataset}: We introduce ActivityNarrated, a semi-naturalistic wearable activity dataset with 22 participants, up to 15 body-worn IMU placements with expert annotations, participant narrations, weak video-language labels, and a compact 23-class taxonomy for closed-set comparison.
    
    \item \textbf{ActNarrator model}: We present ActNarrator, a three-stage dense narration model that tokenizes continuous IMU streams, detects variable-duration events, and adapts event-conditioned sensor tokens to a frozen SLM for open-vocabulary captioning. Experiments show that ActNarrator produces robust dense sensor captions, improves closed-set diagnostic Macro-F1 over SOTA HAR baselines by 3.8--31.6\%, and provides a reusable sensor-language representation for multiple downstream activity-understanding tasks.

    %\item \textbf{Open-ended wearable activity formulation}: We reformulate wearable HAR as dense sensor--language event captioning, where models localize activity intervals in continuous IMU streams and describe them using open-vocabulary natural language rather than limited classification from a predefined label set.
    % experiment and evaluation protocols 

    %\item \textbf{ActivityNarrated dataset and benchmark}: We introduce a semi-naturalistic wearable dataset with 22 participants, up to 15 IMU placements, expert interval descriptions, participant narrations, weak video-language labels, and a 23-class diagnostic taxonomy. The benchmark evaluates temporal localization, open-vocabulary captioning, sensor--language retrieval, closed-set diagnostic classification, held-out activity generalization, timeline question answering, cross-subject transfer, cross-position transfer, and missing-sensor robustness.
    % dataset 

    %\item \textbf{ActNarrator reference model}: We present a language-conditioned architecture for dense wearable activity narration. The model converts continuous IMU streams into discrete motion tokens and jointly predicts activity intervals and open-vocabulary descriptions from heterogeneous, partially observed sensor placements.
    % model - 3-stage sensor-token-event-caption
    % validate with dataset vs sota showing .... as a foundation method - train once solve many problems (finish conclusion framing)
    
\end{itemize}

\section{Related Work}
\label{sec:related_work}

Wearable HAR lies at the intersection of sensing, machine learning, and human behavior. 
We review prior work along three dimensions that motivate ActivityNarrated: modeling and evaluation practices in wearable HAR, representative wearable HAR datasets, and recent language-grounded approaches for activity understanding. 
Across these areas, our goal is to clarify why open-ended wearable activity narration requires changes in the task formulation, data collection protocol, and evaluation methodology.

\subsection{Modeling and Evaluation Practices in Wearable HAR}
\label{sec:related_trends}

Most wearable HAR pipelines rely on three assumptions: continuous behavior is segmented into fixed-length temporal windows, models are trained and evaluated under fixed sensing configurations, and recognition is framed as closed-set classification over a predefined taxonomy \cite{guan2017lstmensemble,abedin2021attend,murahari2018attentionISWC}. 
These assumptions make benchmarking tractable, but they also constrain what can be evaluated. 
Variable-duration behavior, missing or heterogeneous sensor placements, overlapping activities, and semantic variation across people and contexts are often treated as secondary challenges rather than as central properties of the task.
Recent work has relaxed parts of this formulation. 
Self-supervised and contrastive methods learn transferable representations from unlabeled sensor streams \cite{tang2021selfhar,haresamudram2022assessssl,haresamudram2021cpc}. 
Longer-range temporal models and activity-boundary detection methods move beyond isolated window classification toward continuous-stream reasoning \cite{temporalactionlocalization2024}. 
Domain generalization and cross-dataset HAR studies show that models can degrade under shifts in users, devices, environments, and sensing conditions, motivating domain-invariant or broadly transferable representations \cite{wang2020wearingdiversity,hong2024crosshar,dai2024contrastsense,goat2024}. 
Other work addresses partial sensor availability and heterogeneous sensing configurations, reflecting deployment settings where complete and standardized sensor suites are rarely guaranteed \cite{kang2022partial}. 
Complementary studies examine intra-class variability, personal style, and long-tail imbalance, which challenge the assumption that activity classes are stable and mutually exclusive across participants \cite{su2022bpd,kwon2021imutube}.

These directions improve the robustness and flexibility of wearable HAR, but evaluation often remains tied to fixed windows, prescribed labels, or standardized sensing configurations. 
As a result, it remains difficult to assess whether a model captures activity meaning in a way that generalizes across users, durations, and sensor placements, or whether it primarily fits dataset-specific taxonomies and collection protocols. 
ActivityNarrated addresses this gap by making temporal localization, placement variability, missing-sensor robustness, and open-vocabulary activity semantics part of the benchmark itself.

\subsection{Wearable HAR Datasets}
\label{sec:related_datasets}

Early influential datasets such as WISDM/ActiTracker~\cite{kwapisz2011wisdm}, UCI HAR~\cite{anguita2013uci}, and PAMAP2~\cite{reiss2012pamap2} established the standard wearable HAR benchmark pattern: participants perform predefined activities, inertial data are collected from one or a small number of body locations, and models are evaluated using fixed-window classification. 
These datasets enabled foundational progress because they are accessible, standardized, and easy to compare across methods.
Later datasets expanded HAR along several dimensions. 
Opportunity~\cite{chavarriaga2013opportunity} introduced sensor-rich recordings of daily activities with fine-grained and compositional labels. 
RealWorld HAR~\cite{sztyler2016realworld} captured simultaneous recordings from multiple body positions. 
SHL~\cite{gjoreski2018shl} supported large-scale mobile locomotion and transportation recognition. 
Datasets such as ExtraSensory~\cite{vaizman2018extrasensory}, MM-Fit~\cite{stromback2020mmfit}, WEAR~\cite{bock2024wear}, and UTD-MHAD~\cite{chen2015utd} further broadened wearable HAR through in-the-wild collection, multimodal sensing, richer activity labels, or more diverse participant populations. 
Despite this progress, existing datasets typically cover only part of the open-world HAR problem. 
Many rely on scripted or semi-scripted activities, which improves comparability but limits participant-specific and long-tail behavior. 
Many use one or a small number of fixed body placements, making it difficult to evaluate systematic cross-position or missing-sensor generalization. 
Most also use closed vocabularies and mutually exclusive labels, which limits evaluation of unseen activities, overlapping actions, and multiple valid descriptions of the same interval. 
Language-labeled wearable datasets are especially relevant, but they typically do not jointly provide unscripted behavior, dense multi-position IMU sensing, time-aligned natural-language supervision, continuous event localization, and open-vocabulary evaluation.

To our knowledge, no prior wearable HAR benchmark combines these properties at the same scale and within a unified evaluation protocol.
ActivityNarrated is designed around this combination: participants act without a shared activity script, activities are observed across up to 15 body positions, missing sensors are explicitly considered, and supervision is provided through temporally aligned natural-language descriptions together with a compact closed-set taxonomy for diagnostic comparison.

\subsection{Language-Grounded and Open-Vocabulary Wearable Recognition}
\label{sec:related_open_vocab}

Language-supervised representation learning has transformed recognition tasks in vision and multimodal learning. 
Models such as CLIP align perceptual inputs with natural-language descriptions, enabling recognition systems to be queried with arbitrary text and to generalize beyond the exact label sets used during training \cite{radford2021clip}. 
This paradigm has been extended to open-vocabulary detection \cite{minderer2023scalingovod}, segmentation \cite{yuan2024ovsam}, and video understanding \cite{weng2023openvclip}. 
These developments suggest a useful direction for wearable sensing: activity recognition can be treated not only as label prediction, but also as semantic alignment between sensor evidence and language.
Wearable HAR has begun to adopt this perspective. 
Systems such as \textit{Sensor2Text}~\cite{chen2024sensor2text} and \textit{UbiPhysio}~\cite{wang2024ubiphysio} connect wearable signals with natural-language descriptions to support querying, summarization, or feedback. 
Other approaches, including \textit{One Model to Fit Them All}~\cite{wei2025one}, \textit{IMUZero}~\cite{su2025imuzero}, \textit{IMUGPT2}~\cite{leng2024imugpt2}, and LLM-guided semantic alignment methods~\cite{su2025semantic}, use language or LLM-derived semantics to improve transfer, zero-shot recognition, or cross-dataset generalization.
These approaches demonstrate the value of language for wearable sensing, but many still inherit assumptions from closed-set HAR. 
Some derive language supervision from class names, templates, or existing labeled datasets, limiting the diversity of activity descriptions. 
Some operate on fixed-length windows or pre-segmented intervals, avoiding the problem of detecting activity boundaries in continuous streams. 
Others evaluate primarily through known classes, standardized placements, or retrieval settings that do not jointly test temporal localization and open-ended description.

ActivityNarrated differs in both formulation and evaluation. 
Rather than adding language as an auxiliary interface to a closed-set benchmark, it treats time-aligned activity descriptions as the primary supervision and output representation for continuous wearable streams. 
This supports a more scalable and foundational model style pretraining without closed-set fixed-window constraints, which enables wide range of possibilities of downstream utilization from a single model, and encompasses closed-set classification as a zero-shot downstream application, and significantly outperforms state-of-the-art HAR methods in terms of prediction performance, robustness and generalization.

\section{ActNarrator: a Three-Stage Architecture for Dense Sensor--Language Event Captioning}
\label{sec:approach}

We present ActNarrator, a reference architecture for open-ended wearable activity narration from continuous IMU streams. 
Rather than classifying pre-segmented windows into fixed activity labels, ActNarrator takes one or more continuous IMU streams over a long temporal context and predicts a set of activity intervals, each paired with a natural-language description. 
The model is intended as a baseline instantiation of the ActivityNarrated benchmark rather than as the only possible architecture for the task.

ActNarrator has three stages:
\begin{enumerate}
    \item A VQ-VAE-based \textit{tokenizer} is trained to discretize continuous multi-channel IMU streams into sequences of reusable motion tokens to provide a compact representation of local motion patterns. % and reduce the length mismatch between high-frequency sensor streams and language-model inputs.
    \item A self-attention-based \textit{event segmentation module}  predicts the temporal start and end boundaries of events. This allows arbitrary event instances and lengths in a longer context horizon (20 seconds) which is necessary in unscripted activity settings.
    \item A lightweight transformer-based \textit{Q-Former} aligns the sensor-specific tokens with language embeddings to produce semantically aligned sensor tokens, which are then used by a frozen off-the-shelf small language model (SLM) to generate captions for each segmented activity instance or reason about the activities with different prompt-based strategies. 
\end{enumerate}
 
Stage 1 is trained with only the sensor data; while Stage 2 and 3 are trained together, with the natural-language annotations, to ensure coherent temporal segmentation and semantic context.

\begin{figure*}[t]
    \centering
    \includegraphics[width=0.8\linewidth]{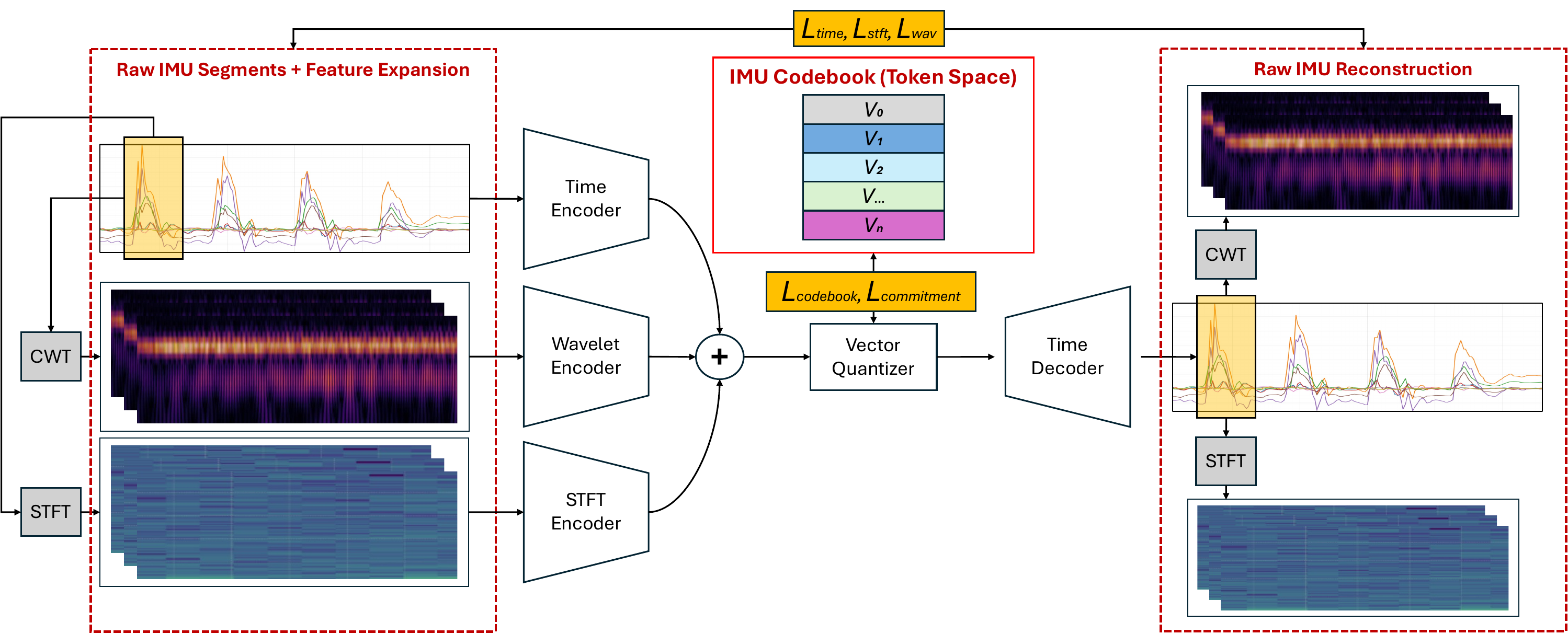}
    \caption{\textbf{Spectral VQ-VAE for IMU tokenization.} Continuous IMU streams are split into overlapping chunks and represented through time-domain, STFT, and wavelet views. The view-specific encoders produce latent embeddings that are fused and quantized through a shared codebook. The decoder reconstructs time-domain and spectral targets during tokenizer training.}
    \label{fig:spectral_vqvae}
\end{figure*}

\subsection{IMU Sensor Signal Tokenization with Spectral VQ-VAE}
\label{sec:tokenization}

IMU streams are high-frequency continuous signals, whereas LLMs operate on compact discrete token sequences. 
For each sensor position, we divide the IMU stream into non-overlapping 2\,s chunks. 
At the dataset sampling rate of approximately 30\,Hz, each chunk contains about 60 samples and produces one discrete motion token.

Each chunk is represented through three complementary views: raw time-domain IMU signals, short-time Fourier transform (STFT) features, and continuous wavelet transform (CWT) features. 
The time-domain view captures local motion amplitude and direction, while the spectral views capture periodic and transient structure that is useful for activities such as walking, jumping, reaching, or object manipulation.

Each view is processed by a lightweight encoder. 
The resulting embeddings are fused and vector-quantized using a shared codebook with $K=128$ entries, producing a discrete token for each IMU chunk. 
The tokenizer is trained as a VQ-VAE using reconstruction losses($\mathcal{L}_{\mathrm{time}}$, $\mathcal{L}_{\mathrm{stft}}$, $\mathcal{L}_{\mathrm{wav}}$), codebook loss ($\mathcal{L}_{\mathrm{codebook}}$), and commitment loss ($\mathcal{L}_{\mathrm{commitment}}$). 
After tokenizer training, each sensor stream is converted into a temporally ordered sequence of discrete motion tokens. 
Formal tokenizer definitions and losses are provided in Appendix~\ref{app:actnarrator_tokenizer}.

\subsection{Dense Event Temporal Localization from Multi-Position IMU Tokens}
\label{sec:generation}

\begin{figure*}[t]
    \centering
    \includegraphics[width=\linewidth]{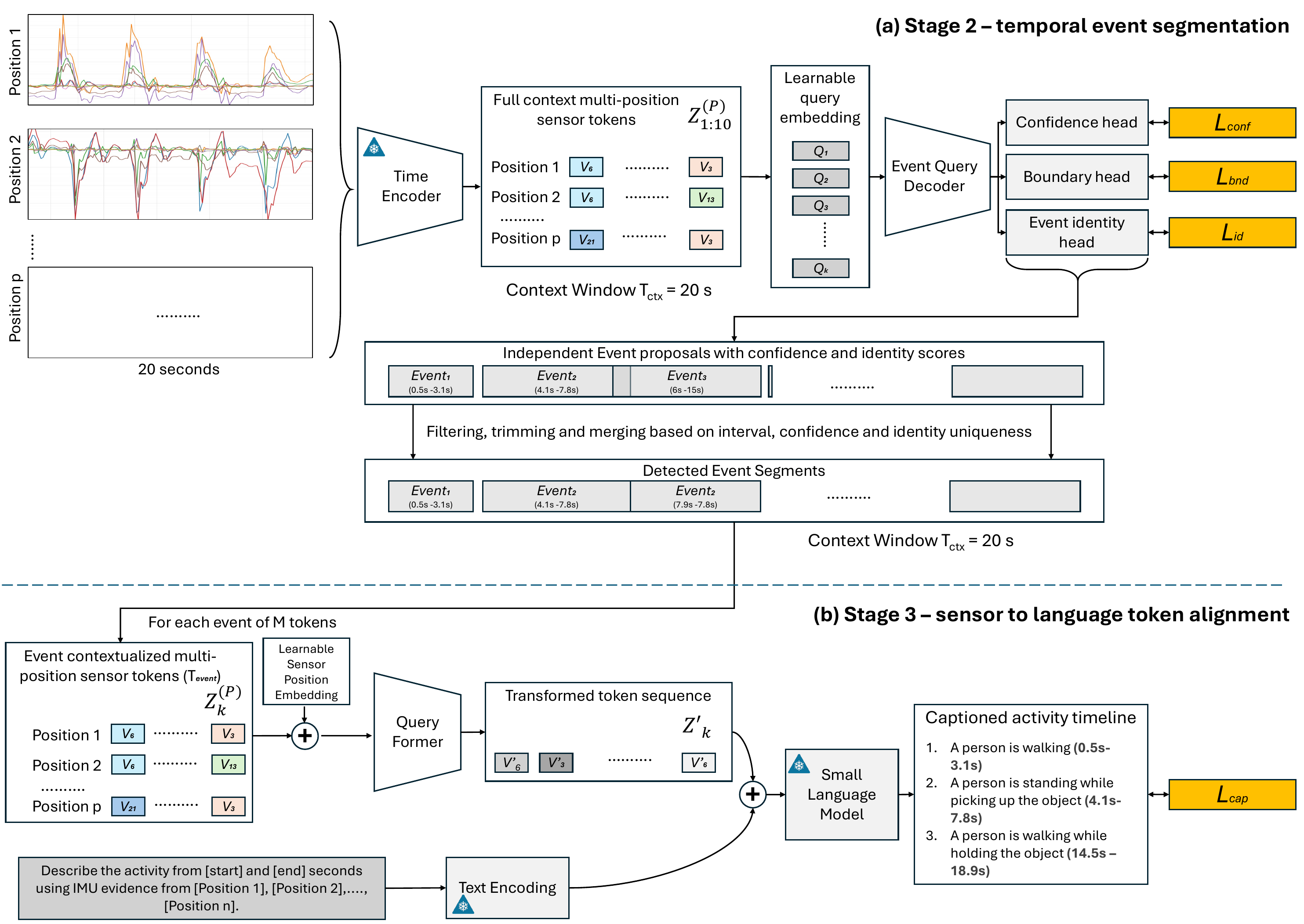}
    \caption{\textbf{Sensor-to-LLM generation.} Token sequences from an arbitrary subset of IMU positions are embedded and fused via a Q-Former into fixed-size sensor prompts. A textual header specifies positions and duration. A frozen LLM generates open-vocabulary activity descriptions conditioned on both prompts and header.}
    \label{fig:sensor_llm}
\end{figure*}

For a context window of duration $T_{\mathrm{ctx}}=20$\,s and chunk size of 2\,s, each observed sensor position $p$ produces a sequence of $M=10$ tokens $Z^{(p)}_{1:M}$. 
An \textit{event segmentation module} is designed to process token sequences per sensor and predict candidate activity events' start and end positions without predicting the semantic descriptions. 
Each predicted event $k$ is represented by a temporal interval
$T_k=[s_k,e_k]$, where $0 \leq s_k < e_k \leq T_{\mathrm{ctx}}$, and by the corresponding subset of sensor tokens
$Z_k \subseteq \{Z^{(p)}_{1:M}\}_{p \in \mathcal{P}_{\mathrm{obs}}}$ that fall inside the predicted interval.

To segment the events, the module uses a fixed set of $K$ learnable event queries. Each query represents a possible activity event within $T_{\mathrm{ctx}}$. The queries are passed through an event query decoder, where they interact with each other through self-attention and attend to the encoded IMU timeline through cross-attention, allowing them to focus on the sensor sequence regions relevant to a candidate event.

The decoded query representations are then passed to temporal proposal heads. A confidence head predicts whether each query corresponds to a real activity event or to no event. A boundary head predicts a normalized temporal center and duration, which are converted into start and end times within $T_{\mathrm{ctx}}$. Thus, although the input tokens are produced every 2\,s, the predicted boundaries are continuous values in $T_{\mathrm{ctx}}$ and are not restricted to 2\,s multiples. For each valid proposal, the predicted interval defines an event-specific token region $Z_k$, which is later used by the captioning module. An additional event identity head produces an identity embedding for each predicted event. This embedding captures the motion pattern and temporal context of the proposal and is used to compare proposals generated from overlapping context windows.

During training, predicted event queries are matched to ground-truth intervals using one-to-one Hungarian matching based on confidence, boundary distance, and temporal overlap. Matched queries are optimized with a confidence loss ($\mathcal{L}_{\mathrm{conf}}$) and a boundary regression loss ($\mathcal{L}_{\mathrm{bnd}}$), encouraging high event confidence and accurate temporal boundaries. Unmatched queries are trained with a no-event loss so that unused queries do not produce false event proposals. The event identity head is supervised with an identity consistency loss ($\mathcal{L}_{\mathrm{id}}$), where proposals matched to the same ground-truth event are encouraged to have similar identity embeddings, while proposals matched to different events are encouraged to remain distinct.

At inference time, the model is applied over overlapping context windows across the full IMU stream. Each window produces up to $K$ event proposals. Each proposal contains an event interval $T_k=[s_k,e_k]$, its event-token subset $Z_k$, a confidence score, and an event identity embedding. Low-confidence proposals are filtered, and temporally redundant proposals across overlapping windows are merged using temporal non-maximum suppression. The event identity embedding is used during this merging step so that proposals with high temporal overlap and similar event identity are treated as duplicate detections. The remaining predictions are sorted by start time to form the final dense activity timeline.

\subsection{Sensor-Conditioned Caption Generation}
\label{sec:caption_generation}

For each predicted event $k$, ActNarrator summarizes the encoded sensor tokens over the segmented temporal span $T_k=[s_k,e_k]$. The tokens that fall inside this interval form the event-specific token set $Z_k$.

The event-specific token sequence $Z_k$ is passed to a lightweight Q-Former, which maps the sensor-token sequence into the embedding space of a frozen decoder-only SLM. 
An open-weight language model is needed since the language embedding space needs to be exposed to the Q-Former, which is not possible with API-accessed commercial models.
The Q-Former preserves the event-token structure and produces a semantic token sequence $Z'_k$ with the same token count as the selected event tokens. Therefore, for an event containing $n$ temporal tokens from each of $p$ observed sensor positions, $Z'_k$ contains $n \times p$ semantic tokens.

Each input token in  $Z_k$ receives a learned sensor-position embedding following the method from VideoLLaMA\cite{zhang2023video}, and the textual prompt explicitly lists the observed positions. The sequence $Z'_k$ is then combined with a short textual header describing the predicted interval and available sensor positions. 
For example: 

\textit{Describe the activity from [$s_k$] and [$e_k$] seconds using IMU evidence from [Position 1], [Position 2],...., [Position n].}

This header makes the temporal interval and sensor availability explicit, supporting evaluation under different sensor-position subsets. The frozen SLM uses the semantic sensor tokens and textual header to generate one open-vocabulary activity caption for the predicted event.

Caption generation is trained with a next-token prediction loss ($\mathcal{L}_{\mathrm{cap}}$) on the natural-language event descriptions ground-truth. This loss conditions the Q-Former and projection layers to convert sensor tokens $Z_k$ to semantic tokens  $Z'_k$  as the input of the frozen SLM to generate activity descriptions that match the ground truth.

\section{Evaluation Methods and Results}
\label{sec:evaluation}

We evaluate ActNarrator along eight axes: IMU tokenization, temporal localization, dense open-vocabulary captioning and retrieval, missing-sensor robustness, closed-set diagnostic classification, external transfer, downstream activity timeline question answering, and held-out activity-class generalization.

\subsection{Training and Evaluation Setup}
\label{sec:train_config}

\subsubsection{Training and validation premise.}
All trainable components are optimized with Adam using a learning rate of $1\times10^{-3}$, weight decay $1\times10^{-4}$, and batch size 64.
Training runs for up to 300 epochs with early stopping patience 10 based on task-specific validation metrics.
The final checkpoint is selected by validation performance.
Experiments are conducted on a Linux system with an NVIDIA A6000 GPU, AMD Ryzen~9 CPU, and PyTorch with CUDA support.
Unless otherwise stated, ActivityNarrated experiments are run under leave-one-subject-out cross validation and reported as mean $\pm$ standard deviation across folds.
All confidence thresholds, early-stopping decisions, and model-selection choices are determined using validation data within each fold.

% \paragraph{Dense event-captioning setup.}
% For continuous-stream inference, we use overlapping context windows of duration $T_{\mathrm{ctx}}=20$\,s.
% The tokenizer uses a 2\,s chunk size, producing one discrete motion token every 2\,s for each available sensor position.
% Thus, each context window contains at most 10 tokens per sensor position before multi-position concatenation.
% Each context window is processed by learnable event queries that jointly predict event confidence, start time, end time, and an event-conditioned caption.
% During training, predicted event queries are matched to annotated intervals using one-to-one assignment based on temporal overlap and boundary quality.
% At inference time, ActNarrator slides the 20\,s context window over the full IMU stream, predicts event-caption proposals within each window, and merges overlapping proposals using confidence thresholding and temporal non-maximum suppression.

\subsubsection{Language model backbones.}
All SLM parameters remain frozen.
Only the temporal encoder, event segmentation module, Q-Former and projection layers are trained.
The main ActNarrator configuration uses Qwen~2.5~7B as the frozen SLM backbone based on ablation validation performance.
Additional frozen backbones, including Qwen~2.5 1.5B/3B, Gemma~2B, and LLaMA~8B variants, are evaluated in Appendix~\ref{app:llm_backbone_scaling}.

\subsubsection{Baseline and adaptation.}
We compare against representative closed-set classification and language-grounded baselines.
For closed-set HAR, DeepConvLSTM~\cite{singh2020deep}, TinyHAR~\cite{zhou2022tinyhar}, and CrossHAR~\cite{hong2024crosshar} are trained with our ActivityNarrated dataset configured with fixed-size rolling windows and predefined hard labels consolidated from the open-ended annotations.
For language-grounded evaluation, IMU2CLIP~\cite{moon2023imu2clip}, OVHAR~\cite{ray2025initial}, %PaLM-E-style sensor prompting~\cite{driess2023palm},
SensorLLM~\cite{li2025sensorllm}, SensorLM~\cite{zhang2025sensorlm}, and HARGPT~\cite{ji2024hargpt} are adapted to use the same train/test splits and normalized text pools as ActNarrator.

\subsubsection{Evaluation metrics.}
Open-vocabulary wearable activity understanding over continuous streams requires evaluation along multiple axes: temporal localization, natural-language caption quality, sensor--language alignment, closed-set comparability, representation quality, downstream timeline reasoning, and held-out activity generalization. Table~\ref{tab:detailed_metric_groups} summarizes the benchmark metrics, the evaluation protocol, and the purpose of each metric group.
Across experiments, we emphasize deployment-relevant generalization under cross-subject (XS), cross-subject-and-position (XSP), missing-sensor, held-out activity-class, and external-dataset conditions.

\begin{table*}[t]
\centering
\scriptsize
\caption{
Detailed benchmark metrics used in ActivityNarrated. Closed-set metrics are included for comparability with conventional HAR, but dense open-vocabulary activity narration is the primary benchmark objective.
}
\label{tab:detailed_metric_groups}
\begin{tabular}{p{0.20\linewidth}p{0.15\linewidth}p{0.34\linewidth}p{0.20\linewidth}}
\toprule
\textbf{Evaluation Dimension} & \textbf{Metrics} & \textbf{Evaluation protocol} & \textbf{What it measures} \\
\midrule

Temporal event segmentation
&
Recall, Mean Average Precision (mAP), boundary error
&
Predicted intervals are confidence-filtered and matched to ground-truth intervals using temporal IoU. Detection metrics are computed after greedy matching at each tIoU threshold. Boundary error is computed on matched predictions.
&
Whether the model detects \emph{when} activity events occur and how accurately the predicted start/end times align with annotations.
\\

\midrule

Caption quality
&
BERTScore
&
Computed only for temporally matched prediction--ground-truth pairs with $\mathrm{tIoU}\geq0.5$.
&
Whether generated captions semantically match valid natural-language descriptions of the activity interval.
\\

\midrule

Sensor--language retrieval
&
Hit Rate (Hit), Mean Reciprocal Rank (MRR) , Normalized Discounted Cumulative Gain (nDCG)
&
For each matched event representation, all unique normalized activity descriptions in the test fold are ranked. The candidate pool is fixed within each fold and shared across methods. A retrieval is considered correct if the similarity score between the prediction and at least one candidate description exceeds $0.5$. 
&
Whether sensor/event representations align with the correct activity descriptions in an open-vocabulary text pool.
\\

\midrule

Closed-set diagnostic classification
&
Accuracy, Macro-F1
&
Activity intervals are mapped to a 23-class movement-centric taxonomy. Generated captions are mapped to the same taxonomy using a fixed controlled prompt with deterministic decoding.
&
Compatibility with conventional HAR evaluation, while treating closed-set classification as a diagnostic rather than the primary task.
\\

\midrule

Representation diagnostics
&
Time Reconstruction Error (Time $\ell_1$), Spectral Reconstruction Error (spectral $\ell_1$), Jensen Shannon Divergence (JS divergence)
&
Applied to token-based models. Reconstruction errors compare recovered IMU signals or spectral features against the input. JS divergence measures token-distribution stability across subjects, positions, or missing-sensor settings.
&
Whether learned motion tokens preserve useful IMU structure and remain stable under subject, placement, or sensor-availability shifts.
\\

\midrule

Timeline question answering
&
Temporal ordering QA accuracy, Time localization QA accuracy, Counting QA Accuracy
semantic QA accuracy
&
Questions are constructed over temporally ordered event timelines. Taxonomy-compatible questions use class labels and timestamps; semantic questions require natural-language event descriptions.
&
Whether predicted activity timelines support downstream reasoning about what happened, when it happened, and in what order.
\\

\midrule

Held-out activity-class generalization
&
Held out class Accuracy, Held out Macro F1
&
Selected activity classes are removed from paired sensor-label and sensor-text training examples. At inference time, class names are introduced only through text candidates or caption-to-class mapping.
&
Whether open-vocabulary methods can recognize activity classes whose paired sensor examples are absent during training.
\\

\bottomrule
\end{tabular}
\end{table*}

Unless otherwise specified, interval-level metrics are computed after matching predicted events to ground-truth events. Predicted intervals are filtered using a validation-selected confidence threshold. Unmatched predictions are treated as false positives for detection-style metrics, and unmatched ground-truth intervals are treated as missed events. All interval-level results are averaged over evaluation folds and reported as mean $\pm$ standard deviation. Full metric definitions are provided in Appendix~\ref{app:metric_definitions}.

\subsection{IMU Tokenization Quality and Stability}
\label{sec:imu_disc}

We first evaluate the Spectral VQ-VAE tokenizer in isolation.
The tokenizer should preserve time-domain and spectral signal structure while producing reusable motion tokens across subjects and sensor positions.
We evaluate reconstruction quality using time-domain and spectral $\ell_1$ error, and token stability using Jensen--Shannon divergence across subjects and positions.

Table~\ref{tab:disc_compare_main} compares raw time-domain, STFT-only, wavelet-only, and combined spectral tokenizers.
The combined Spectral VQ-VAE achieves the lowest reconstruction error and token divergence under both XS and XSP settings.
Under XSP, it reduces time-domain error from $0.381$ for the raw VQ-VAE to $0.095$, and reduces JS divergence from $0.59$ to $0.24$.
These results suggest that jointly encoding time-domain, STFT, and wavelet views improves both signal fidelity and token stability under subject and placement shift.
Additional ablations over dictionary size, chunk length, and qualitative token reuse are provided in Appendix~\ref{app:tokenizer_ablations}.

\begin{table}[h]
\caption{
Comparison of IMU tokenization strategies on ActivityNarrated.
Lower reconstruction error indicates better signal fidelity; lower JS divergence indicates more stable token usage across subjects or positions.
}
\centering
\scriptsize
\setlength{\tabcolsep}{5pt}
\begin{tabular}{l l ccc}
\toprule
\textbf{Split}
& \textbf{Tokenizer}
& \textbf{Time $\ell_1$} $\downarrow$
& \textbf{Spectral $\ell_1$} $\downarrow$
& \textbf{JS} $\downarrow$ \\
\midrule
XS & Raw VQ-VAE & 0.332 $\pm$ 0.018 & 0.471 $\pm$ 0.026 & 0.54 $\pm$ 0.04 \\
& STFT VQ-VAE & 0.148 $\pm$ 0.010 & 0.213 $\pm$ 0.014 & 0.46 $\pm$ 0.05 \\
& Wavelet VQ-VAE & 0.156 $\pm$ 0.011 & 0.198 $\pm$ 0.013 & 0.39 $\pm$ 0.05 \\
& \textbf{Spectral VQ-VAE} & \textbf{0.082 $\pm$ 0.004} & \textbf{0.116 $\pm$ 0.006} & \textbf{0.21 $\pm$ 0.02} \\
\midrule
XSP & Raw VQ-VAE & 0.381 $\pm$ 0.022 & 0.512 $\pm$ 0.029 & 0.59 $\pm$ 0.03 \\
& STFT VQ-VAE & 0.169 $\pm$ 0.012 & 0.241 $\pm$ 0.016 & 0.51 $\pm$ 0.05 \\
& Wavelet VQ-VAE & 0.178 $\pm$ 0.013 & 0.226 $\pm$ 0.015 & 0.44 $\pm$ 0.05 \\
& \textbf{Spectral VQ-VAE} & \textbf{0.095 $\pm$ 0.006} & \textbf{0.131 $\pm$ 0.008} & \textbf{0.24 $\pm$ 0.02} \\
\bottomrule
\end{tabular}
\label{tab:disc_compare_main}
\end{table}

\subsection{Temporal Event Segmentation Performance}
\label{sec:temp_local}

We next evaluate whether ActNarrator can localize activity intervals in continuous IMU streams without oracle segmentation at inference time.
Models must predict candidate intervals, assign confidence scores, and align these predictions with annotated activity segments.

\begin{table*}[h]
\caption{
Temporal localization performance on ActivityNarrated.
Higher Recall@$\tau$ and mAP indicate better event detection; lower boundary error indicates more precise interval localization.
}
\centering
\scriptsize
\setlength{\tabcolsep}{6pt}
\begin{tabular}{l cccccc}
\toprule
\textbf{Split}
&
\textbf{Method}
& \textbf{Rec@0.3} $\uparrow$
& \textbf{Rec@0.5} $\uparrow$
& \textbf{Rec@0.7} $\uparrow$
& \textbf{mAP} $\uparrow$
& \textbf{Bnd. Err.} $\downarrow$ \\
\midrule
XS & Raw + Q-Former & 62.5 $\pm$ 1.7 & 49.3 $\pm$ 1.6 & 31.7 $\pm$ 1.4 & 39.5 $\pm$ 1.5 & 1.48 $\pm$ 0.10 \\
& Spectral + mean pooling & 64.1 $\pm$ 1.6 & 51.0 $\pm$ 1.5 & 33.2 $\pm$ 1.4 & 41.3 $\pm$ 1.4 & 1.39 $\pm$ 0.09 \\
& Spectral + attention & 68.6 $\pm$ 1.6 & 55.7 $\pm$ 1.5 & 37.9 $\pm$ 1.3 & 46.1 $\pm$ 1.4 & 1.20 $\pm$ 0.08 \\
& Spectral + Q-Former & 74.2 $\pm$ 1.4 & 61.8 $\pm$ 1.4 & 42.6 $\pm$ 1.2 & 52.5 $\pm$ 1.3 & 1.02 $\pm$ 0.07 \\
& \textbf{ActNarrator} & \textbf{78.9 $\pm$ 1.3} & \textbf{66.4 $\pm$ 1.3} & \textbf{46.8 $\pm$ 1.2} & \textbf{57.1 $\pm$ 1.2} & \textbf{0.88 $\pm$ 0.06} \\
\midrule
XSP & Raw + Q-Former & 55.8 $\pm$ 1.9 & 42.6 $\pm$ 1.7 & 26.3 $\pm$ 1.5 & 34.0 $\pm$ 1.5 & 1.75 $\pm$ 0.12 \\
& Spectral + mean pooling & 57.0 $\pm$ 1.8 & 44.3 $\pm$ 1.7 & 27.9 $\pm$ 1.4 & 35.7 $\pm$ 1.5 & 1.65 $\pm$ 0.11 \\
& Spectral + attention & 61.5 $\pm$ 1.7 & 48.8 $\pm$ 1.6 & 32.4 $\pm$ 1.4 & 40.5 $\pm$ 1.4 & 1.42 $\pm$ 0.10 \\
& Spectral + Q-Former & 68.0 $\pm$ 1.6 & 55.1 $\pm$ 1.5 & 37.8 $\pm$ 1.3 & 47.2 $\pm$ 1.4 & 1.19 $\pm$ 0.08 \\
& \textbf{ActNarrator} & \textbf{72.6 $\pm$ 1.5} & \textbf{59.7 $\pm$ 1.4} & \textbf{41.3 $\pm$ 1.3} & \textbf{51.8 $\pm$ 1.3} & \textbf{1.05 $\pm$ 0.07} \\
\bottomrule
\end{tabular}
\label{tab:temporal_main}
\end{table*}

Table~\ref{tab:temporal_main} reports temporal event segmentation performance under XS and XSP.
Under XS, \textit{Spectral + Q-Former} reaches Recall@0.5 of $61.8$ and mAP of $52.5$, compared with $49.3$ and $39.5$ for \textit{Raw + Q-Former}.
The full ActNarrator model further improves Recall@0.5 to $66.4$ and mAP to $57.1$.

Under XSP, all methods degrade, confirming that unseen sensor positions remain challenging.
However, ActNarrator retains the strongest performance, achieving Recall@0.5 of $59.7$, mAP of $51.8$, and boundary error of $1.05$\,s.
These results show that spectral motion tokens improve both semantic representation and event-boundary prediction in continuous wearable streams.

\subsection{Dense Open-Vocabulary Captioning Quality }
\label{sec:evaluation_open_vocab}

We evaluate open-vocabulary wearable understanding as dense sensor--language event captioning.
Given continuous IMU streams from available body-worn sensors, the model predicts activity intervals and generates natural-language descriptions for each predicted event.
After temporal matching, generated captions are evaluated with BERTScore, and representation alignment is evaluated using Hit@1, Hit@5, MRR, and nDCG@5.

To test how well the distribution of the generated captions fall within the given ground-truth set, we use a retrieval-style evaluation method from a candidate-pool.
Each test fold uses a fixed candidate text pool consisting of all unique normalized activity descriptions in the test split, without negative subsampling.
For ActivityNarrated, the average candidate-pool size is 151 normalized descriptions.
All methods are evaluated against the same candidate pool within each fold.

Table~\ref{tab:semantic_main} reports the main open-vocabulary results.
Under XS, ActNarrator achieves $36.2$ Hit@1, $68.4$ Hit@5, $0.55$ MRR, and $0.65$ nDCG@5.
The strongest prior baseline, SensorLM, reaches $24.7$ Hit@1 and $0.53$ nDCG@5.
Under XSP, ActNarrator achieves $32.5$ Hit@1 and $0.61$ nDCG@5, compared with $22.0$ Hit@1 and $0.50$ nDCG@5 for SensorLM.
These results suggest that dense localization, spectral tokenization, and event-conditioned language prompting jointly improve sensor--language grounding under both subject and placement shift.
The full details of quantitative captioning evaluation are provided in Appendix \ref{app:metrics_caption_retrieval}

\begin{table*}[h]
\caption{
Dense open-vocabulary event captioning performance on ActivityNarrated.
Metrics are computed on temporally matched predicted intervals.
Hit@K, MRR, and nDCG@5 use fixed test-fold candidate pools with an average size of 151 normalized descriptions.
}
\centering
\scriptsize
\setlength{\tabcolsep}{5pt}
\begin{tabular}{l cccccc}
\toprule
\textbf{split} &
\textbf{Method}
& \textbf{BERTScore} $\uparrow$
& \textbf{Hit@1} $\uparrow$
& \textbf{Hit@5} $\uparrow$
& \textbf{MRR} $\uparrow$
& \textbf{nDCG@5} $\uparrow$ \\
\midrule
XS & IMU2CLIP & 0.514 $\pm$ 0.012 & 9.8 $\pm$ 0.6 & 31.4 $\pm$ 1.3 & 0.21 $\pm$ 0.02 & 0.33 $\pm$ 0.03 \\
& OVHAR & 0.583 $\pm$ 0.011 & 20.6 $\pm$ 0.8 & 45.9 $\pm$ 1.5 & 0.35 $\pm$ 0.03 & 0.49 $\pm$ 0.02 \\
& SensorLLM & 0.612 $\pm$ 0.011 & 23.0 $\pm$ 0.8 & 50.1 $\pm$ 1.6 & 0.39 $\pm$ 0.03 & 0.51 $\pm$ 0.03 \\
& SensorLM & 0.631 $\pm$ 0.010 & 24.7 $\pm$ 0.9 & 52.8 $\pm$ 1.7 & 0.41 $\pm$ 0.03 & 0.53 $\pm$ 0.03 \\
& HARGPT & 0.557 $\pm$ 0.013 & 16.9 $\pm$ 0.7 & 40.2 $\pm$ 1.4 & 0.31 $\pm$ 0.03 & 0.44 $\pm$ 0.03 \\
& Spectral + Q-Former & 0.701 $\pm$ 0.009 & 32.8 $\pm$ 1.0 & 63.5 $\pm$ 1.8 & 0.50 $\pm$ 0.04 & 0.60 $\pm$ 0.03 \\
& \textbf{ActNarrator} & \textbf{0.734 $\pm$ 0.008} & \textbf{36.2 $\pm$ 1.1} & \textbf{68.4 $\pm$ 1.9} & \textbf{0.55 $\pm$ 0.03} & \textbf{0.65 $\pm$ 0.04} \\
\midrule
XSP & IMU2CLIP & 0.489 $\pm$ 0.013 & 8.1 $\pm$ 0.5 & 28.1 $\pm$ 1.3 & 0.19 $\pm$ 0.02 & 0.30 $\pm$ 0.02 \\
& OVHAR & 0.561 $\pm$ 0.012 & 18.0 $\pm$ 0.7 & 42.0 $\pm$ 1.5 & 0.32 $\pm$ 0.02 & 0.46 $\pm$ 0.03 \\
& SensorLLM & 0.588 $\pm$ 0.012 & 20.3 $\pm$ 0.8 & 44.6 $\pm$ 1.5 & 0.36 $\pm$ 0.03 & 0.48 $\pm$ 0.03 \\
& SensorLM & 0.607 $\pm$ 0.011 & 22.0 $\pm$ 0.8 & 47.5 $\pm$ 1.6 & 0.38 $\pm$ 0.03 & 0.50 $\pm$ 0.03 \\
& HARGPT & 0.532 $\pm$ 0.014 & 14.8 $\pm$ 0.7 & 36.2 $\pm$ 1.4 & 0.28 $\pm$ 0.03 & 0.40 $\pm$ 0.03 \\
& Spectral + Q-Former & 0.681 $\pm$ 0.010 & 29.4 $\pm$ 0.9 & 58.7 $\pm$ 1.8 & 0.47 $\pm$ 0.03 & 0.56 $\pm$ 0.03 \\
& \textbf{ActNarrator} & \textbf{0.712 $\pm$ 0.009} & \textbf{32.5 $\pm$ 1.0} & \textbf{63.1 $\pm$ 1.9} & \textbf{0.52 $\pm$ 0.03} & \textbf{0.61 $\pm$ 0.04} \\
\bottomrule
\end{tabular}
\label{tab:semantic_main}
\end{table*}

\subsection{Missing-Sensor Robustness}
\label{sec:missing_sensor}

Real-world wearable deployments rarely provide a complete or fixed sensor configuration.
Sensors may be absent due to hardware failure, user preference, battery limitations, or application-specific constraints.
We therefore evaluate missing-sensor robustness by restricting the available sensor positions at inference time.

We compare two training conditions.
In the first, the model is trained using all 15 positions and evaluated with only a subset available.
This reflects unexpected sensor dropout or partial deployment.
In the second, the model is trained only on the same subset used at inference time, which tests whether specializing to the subset is preferable to learning from broader placement diversity.

\begin{table}[h]
\caption{
Dense open-vocabulary event captioning robustness under missing-sensor evaluation on ActivityNarrated under the XS split.
Metrics are reported on temporally matched predicted intervals after continuous-stream inference.
Hit@K, MRR, and nDCG@5 are computed against the same fixed candidate pool used in the main ActivityNarrated XS retrieval evaluation, with an average candidate-pool size of 151 normalized descriptions.
}
\centering
\scriptsize
\setlength{\tabcolsep}{4pt}
\begin{tabular}{l cccc cccc}
\toprule
& \multicolumn{4}{c}{\textbf{Train on all 15 positions}}
& \multicolumn{4}{c}{\textbf{Train on inference subset}} \\
\cmidrule(lr){2-5} \cmidrule(lr){6-9}
\textbf{Inference sensors}
& \textbf{Hit@1 $\uparrow$}
& \textbf{Hit@5 $\uparrow$}
& \textbf{MRR $\uparrow$}
& \textbf{nDCG@5 $\uparrow$}
& \textbf{Hit@1 $\uparrow$}
& \textbf{Hit@5 $\uparrow$}
& \textbf{MRR $\uparrow$}
& \textbf{nDCG@5 $\uparrow$} \\
\midrule
All 15 positions
& 36.2 $\pm$ 1.1 & 68.4 $\pm$ 1.9 & 0.55 $\pm$ 0.03 & 0.65 $\pm$ 0.04
& 36.2 $\pm$ 1.1 & 68.4 $\pm$ 1.9 & 0.55 $\pm$ 0.03 & 0.65 $\pm$ 0.04 \\
Wrist (r)
& \textbf{29.8 $\pm$ 1.3} & \textbf{57.6 $\pm$ 1.9} & \textbf{0.46 $\pm$ 0.03} & \textbf{0.54 $\pm$ 0.03}
& 17.2 $\pm$ 1.5 & 36.4 $\pm$ 2.2 & 0.26 $\pm$ 0.04 & 0.31 $\pm$ 0.03 \\
Thigh (l)
& \textbf{26.8 $\pm$ 1.2} & \textbf{53.4 $\pm$ 1.8} & \textbf{0.42 $\pm$ 0.03} & \textbf{0.50 $\pm$ 0.02}
& 15.7 $\pm$ 1.4 & 33.5 $\pm$ 2.1 & 0.24 $\pm$ 0.03 & 0.29 $\pm$ 0.03 \\
Head
& \textbf{22.2 $\pm$ 1.4} & \textbf{46.2 $\pm$ 2.1} & \textbf{0.36 $\pm$ 0.03} & \textbf{0.42 $\pm$ 0.02}
& 11.5 $\pm$ 1.3 & 27.2 $\pm$ 2.0 & 0.17 $\pm$ 0.03 & 0.21 $\pm$ 0.03 \\
Head + Wrist (r)
& \textbf{28.5 $\pm$ 1.3} & \textbf{55.9 $\pm$ 1.8} & \textbf{0.44 $\pm$ 0.03} & \textbf{0.52 $\pm$ 0.03}
& 19.1 $\pm$ 1.4 & 38.6 $\pm$ 2.0 & 0.29 $\pm$ 0.03 & 0.34 $\pm$ 0.04 \\
Thigh (l) + Wrist (r)
& \textbf{30.4 $\pm$ 1.2} & \textbf{59.1 $\pm$ 1.9} & \textbf{0.47 $\pm$ 0.03} & \textbf{0.55 $\pm$ 0.03}
& 20.5 $\pm$ 1.5 & 41.2 $\pm$ 2.1 & 0.31 $\pm$ 0.04 & 0.36 $\pm$ 0.03 \\
Head + Thigh (l) + Wrist (r)
& \textbf{31.8 $\pm$ 1.3} & \textbf{61.5 $\pm$ 1.8} & \textbf{0.49 $\pm$ 0.03} & \textbf{0.58 $\pm$ 0.03}
& 22.9 $\pm$ 1.4 & 45.0 $\pm$ 2.0 & 0.34 $\pm$ 0.04 & 0.39 $\pm$ 0.03 \\
\bottomrule
\end{tabular}
\label{tab:ov_missing}
\end{table}

Table~\ref{tab:ov_missing} shows that models trained on all 15 positions degrade gracefully when evaluated with fewer sensors.
With only a single right-wrist IMU, the model retains $29.8$ Hit@1 and $0.54$ nDCG@5.
With thigh and wrist sensors, performance increases to $30.4$ Hit@1 and $0.55$ nDCG@5.
With head, thigh, and wrist sensors, performance reaches $31.8$ Hit@1 and $0.58$ nDCG@5.

Training only on the inference subset performs substantially worse.
For example, wrist-only training achieves $17.2$ Hit@1 and $0.31$ nDCG@5, compared with $29.8$ and $0.54$ when the model is trained on all positions.
This indicates that exposure to diverse sensor placements during training improves partial-observation robustness, even when many positions are missing at test time.
Closed-set missing-sensor diagnostics show the same trend.

\subsection{Closed-Set Classification Comparison with Mainstream HAR Models}
\label{sec:evaluation_closed_set}

Althouggh the primary task is open-vocabulary activity narration, we also evaluate closed-set classification for compatibility with conventional HAR benchmarks.
We use the expert-defined 23-class taxonomy consisting of 22 action roots and one \textit{Other} class.
This evaluation should be interpreted as a downstream diagnostic: it measures how well predicted open-vocabulary intervals and captions can be collapsed into a fixed label space, not the full semantic quality of the generated narratives and the descriptive capabilities.

For ActNarrator, activity events are captioned with natural language, and then mapped to the 23-class taxonomy using a controlled prompt: 
\textit{"Can you classify it to one activity out of these 22 "adjust, bend,...". Classify it as "other" if it does not match any of these."}
This essentially renders closed-set classification as a purely text-level downstream reasoning task.

Traditional HAR baselines operate on fixed rolling windows, and event-level predictions are obtained by aggregating via majority-voting window predictions over the corresponding annotated interval.

Table~\ref{tab:closed_main} reports closed-set classification results.
Under XS, the best fixed-window TinyHAR variant reaches $56.1\%$ accuracy and $34.0\%$ Macro-F1, while CrossHAR reaches $61.8\%$ accuracy and $39.7\%$ Macro-F1.
ActNarrator achieves $80.9\%$ accuracy and $60.8\%$ Macro-F1.
Under XSP, performance drops for all methods, confirming that unseen sensor positions remain challenging.
The full model remains strongest, with $64.0\%$ accuracy and $48.0\%$ Macro-F1.

%These results show that the spectral token representation improves not only open-vocabulary retrieval and captioning, but also conventional fixed-label classification when predictions are collapsed into the 23-class taxonomy.
%However, the persistent gap between XS and XSP confirms that cross-position generalization remains a difficult deployment-relevant challenge.

\begin{table}[h]
\caption{
Closed-set classification on ActivityNarrated.
Macro-F1 is reported in percentage points.
}
\centering
\scriptsize
\setlength{\tabcolsep}{6pt}
\begin{tabular}{l ccc}
\toprule
\textbf{Split}
&
\textbf{Method}
& \textbf{Acc. $\uparrow$}
& \textbf{Macro-F1 $\uparrow$} \\
\midrule
XS & DeepConvLSTM (1 sec) & 48.5 $\pm$ 1.3 & 26.0 $\pm$ 2.4 \\
& DeepConvLSTM (2 sec)  & 51.2 $\pm$ 1.4 & 29.1 $\pm$ 2.0 \\
& DeepConvLSTM (5 sec) & 54.3 $\pm$ 1.5 & 31.4 $\pm$ 2.2 \\
& TinyHAR (1 sec) & 50.1 $\pm$ 1.4 & 28.3 $\pm$ 1.9 \\
& TinyHAR (2 sec)  & 53.0 $\pm$ 1.8 & 31.2 $\pm$ 2.1 \\
 & TinyHAR (5 sec) & 56.1 $\pm$ 1.6 & 34.0 $\pm$ 2.2 \\
 & CrossHAR (1 sec) & 55.1 $\pm$ 1.9 & 35.5 $\pm$ 1.6\\
& CrossHAR (2 sec)  & 57.8 $\pm$ 1.4 & 37.3 $\pm$ 2.5\\
 & CrossHAR (5 sec)& 61.8 $\pm$ 1.5 & 39.7 $\pm$ 2.3 \\
 & Raw features + Q-Former (Dynamic) & 64.2 $\pm$ 1.7 & 43.0 $\pm$ 2.7 \\
 & Spectral tokens + Q-Former (Dynamic) & 77.8 $\pm$ 1.4 & 56.7 $\pm$ 2.4 \\
 & \textbf{ActNarrator (Dynamic)} & \textbf{80.9 $\pm$ 1.3} & \textbf{60.8 $\pm$ 2.2} \\
\midrule
XSP & DeepConvLSTM (1 sec) & 28.0 $\pm$ 1.9 & 14.3 $\pm$ 1.6 \\
& DeepConvLSTM (2 sec)  & 30.5 $\pm$ 1.7 & 15.2 $\pm$ 1.8 \\
 & DeepConvLSTM (5 sec) & 32.8 $\pm$ 1.8 & 16.4 $\pm$ 2.0 \\
& TinyHAR (1 sec) & 30.1 $\pm$ 1.9 & 16.1 $\pm$ 2.0 \\
& TinyHAR (2 sec)  & 33.2 $\pm$ 1.5 & 18.0 $\pm$ 1.8 \\
& TinyHAR (5 sec) & 35.2 $\pm$ 1.9 & 19.3 $\pm$ 2.1 \\
 & CrossHAR (1 sec) & 34.1 $\pm$ 1.5 & 18.8 $\pm$ 1.6 \\
& CrossHAR (2 sec)  & 37.8 $\pm$ 2.1 & 23.5 $\pm$ 1.5 \\
& CrossHAR & 43.6 $\pm$ 1.8 & 28.7 $\pm$ 2.4 \\
& Raw features + Q-Former (Dynamic) & 46.1 $\pm$ 2.0 & 31.0 $\pm$ 2.6 \\
& Spectral tokens + Q-Former (Dynamic) & 60.5 $\pm$ 1.8 & 44.2 $\pm$ 2.5 \\
& \textbf{ActNarrator (Dynamic)} & \textbf{64.0 $\pm$ 1.7} & \textbf{48.0 $\pm$ 2.3} \\
\bottomrule
\end{tabular}
\label{tab:closed_main}
\end{table}

\subsection{Validation on External Open-Volcabulary Dataset}
\label{sec:openmarcie_transfer}

We evaluate our method on OpenMarcie~\cite{bello2026openmarcie} dataset, which is a new IMU HAR dataset both open-vocabulary narration and closed-set labels, where the participants are more constrained to a semi-instructed specialized assembly task (i.e. mostly stationary with upper body movements).

Table~\ref{tab:openmarcie_closed_app} reports closed-set diagnostic classification under the XS setting. We use the same IMU preprocessing, tokenization, and text-normalization pipeline as in the main experiments. The results follow the same trend observed on ActivityNarrated: raw-feature Q-Former improves over standard HAR baselines, spectral tokenization provides a further gain, and the full ActNarrator model performs best. ActNarrator achieves $9 2.5$ accuracy and $90.3$ Macro-F1, outperforming the strongest conventional baseline, CrossHAR, by $4.6$ points in accuracy and $5.2$ points in Macro-F1.

These results suggest that the main architectural advantages of ActNarrator are not specific to the ActivityNarrated dataset, we provide the full retrieval, localization, and additional OpenMarcie analyses in Appendix~\ref{app:openmarcie_detailed}.

\begin{table}[h]
\caption{
OpenMarcie closed-set diagnostic classification under the XS setting.
Macro-F1 is reported in percentage points.
}
\centering
\scriptsize
\setlength{\tabcolsep}{6pt}
\begin{tabular}{l cc}
\toprule
\textbf{Method}
& \textbf{Acc. $\uparrow$}
& \textbf{Macro-F1 $\uparrow$} \\
\midrule
DeepConvLSTM (1 sec) & 85.6 $\pm$ 0.9 & 83.4 $\pm$ 0.7 \\
DeepConvLSTM (2 sec) & 84.9 $\pm$ 1.0 & 82.5 $\pm$ 0.9 \\
DeepConvLSTM (5 sec) & 83.4 $\pm$ 1.1 & 80.7 $\pm$ 1.0 \\
TinyHAR (1 sec) & 85.0 $\pm$ 1.0 & 82.1 $\pm$ 0.9 \\
TinyHAR (2 sec) & 85.8 $\pm$ 1.0 & 83.0 $\pm$ 0.9 \\
TinyHAR (5 sec) & 84.7 $\pm$ 1.1 & 81.8 $\pm$ 1.0 \\
CrossHAR & 87.9 $\pm$ 0.9 & 85.1 $\pm$ 0.8 \\
Raw features + Q-Former (Dynamic) & 88.8 $\pm$ 1.0 & 86.4 $\pm$ 1.0 \\
Spectral tokens + Q-Former (Dynamic) & 91.2 $\pm$ 0.8 & 88.9 $\pm$ 0.8 \\
\textbf{ActNarrator (Dynamic)} & 92.5 $\pm$ 0.7 & 90.3 $\pm$ 0.7 \\
\bottomrule
\end{tabular}
\label{tab:openmarcie_closed_app}
\end{table}

\subsection{Downstream Activity Timeline Question Answering}
\label{sec:timeline_qa}
To evaluate whether open-ended wearable narration enables downstream capabilities beyond conventional closed-set HAR, we introduce an activity timeline question-answering task. 
For each test session, we first construct a one-hour activity horizon: a temporally ordered timeline describing the participant's activities during the hour. 
The task then asks whether model-generated activity timeline can support natural-language queries about what occurred, when it occurred, how often it occurred, and how events were temporally related. 

For each test session, we construct a ground-truth event timeline from human annotations and generate 200 template-based QA pairs covering temporal ordering, time localization, counting, and semantic activity search. 
Example questions include: ``What did the participant do after sitting down?'', ``When did the participant start exercising?'', ``How many drinking-related events occurred?'', and ``Did the participant interact with food?''.
These questions require both temporal grounding and activity semantics; additional details are provided in Appendix~\ref{app:timeline_qa_eval_details}.

We evaluate three timeline sources: ground-truth timelines with human intervals and captions, ActNarrator timelines with predicted intervals and generated captions, and closed-set HAR timelines produced by applying windowed classifiers, merging adjacent windows with the same predicted class, and labeling each merged interval by its class name.
Closed-set timelines provide only coarse class names and timestamps, and therefore cannot represent many object-level, goal-level, or compositional facts.

All sources are evaluated with the same downstream QA procedure.
For each question, we retrieve relevant timeline events using sentence-embedding similarity and provide them to an LLM.
The prompt restricts the model to the provided timeline and requires it to answer ``unknown'' when evidence is insufficient, isolating the effect of the timeline representation.

\begin{table}[htbp]
\centering
\scriptsize
\caption{
Downstream activity timeline QA performance split by question type.
Taxonomy-compatible questions can be answered from coarse class labels and timestamps, while open-vocabulary semantic questions require object-level, goal-level, or compositional activity descriptions.
}
\label{tab:timeline_qa}
\begin{tabular}{lcccc}
\toprule
\textbf{Timeline source} 
& \textbf{Temporal Ordering QA Acc. $\uparrow$} 
& \textbf{Time Localization QA Acc. $\uparrow$} 
& \textbf{Counting QA Acc. $\uparrow$} 
& \textbf{Semantic QA Acc. $\uparrow$} \\
\midrule
Closed-set DeepConvLSTM          
& $62.4 \pm 3.2$
& $67.1 \pm 3.0$
& $59.3 \pm 3.5$
& N/A \\
Closed-set TinyHAR               
& $63.8 \pm 3.0$
& $68.6 \pm 2.8$
& $58.7 \pm 3.5 $
& N/A \\
Closed-set CrossHAR              
& $68.9 \pm 2.7$
& $73.4 \pm 2.5 $
& $60.7 \pm 3.3 $
& N/A \\
ActNarrator                
& $\mathbf{84.7 \pm 2.1}$
& $\mathbf{88.3 \pm 1.9} $
& $ \mathbf{81.5 \pm 2.3} $
& $\mathbf{69.5 \pm 1.8}$ \\
\midrule
Ground-truth captions            
& $100.0 \pm 0.0 $
& $100.0 \pm 0.0 $
& $93.5 \pm 1.7 $
& $74.4 \pm 6.2$ \\
\bottomrule
\end{tabular}%
\end{table}

Table~\ref{tab:timeline_qa} shows that closed-set timelines retain some utility for taxonomy-compatible questions, particularly when the query only requires coarse movement labels and approximate event order.
However, semantic QA is not applicable to closed-set timelines because the required object-level, goal-level, and compositional information is absent from the class-name timeline.

In contrast, ActNarrator timelines support all QA types because they retain temporal intervals together with natural-language event descriptions.
The remaining gap between ActNarrator and ground-truth timelines reflects errors from imperfect event localization and caption generation, while the gap between ActNarrator and closed-set baselines demonstrates a concrete downstream benefit of the proposed open-ended narrative formulation.

\subsection{Classification on Unseen Activities}
\label{sec:heldout_activity_closedset}

To probe the generalization capability on unseen activities, we selectively remove activities from the training dataset, and test the classification performance on those removed activities using conventional HAR metrics.
%This setting tests whether open-vocabulary sensor-language models can recover the correct activity class when paired sensor examples from that class are removed during training.
%Unlike standard closed-set HAR, the evaluated activity classes are not observed as supervised sensor-label or sensor-text examples during training.
We construct the split using the 22-class movement-centric taxonomy derived from expert annotations.
The \textit{Other} class is excluded from the held-out set because it does not correspond to a single coherent activity category.
This experiment is conducted under the cross-subject (XS) protocol.
%For each XS fold, the test participant is held out as in the main cross-subject evaluation.
We hold out three canonical activity classes: \textit{walking}, \textit{sitting}, and \textit{jumping} which were chosen because they span locomotion, posture, and dynamic whole-body motion while sufficiently occur across all participants's recordings.
All events from these classes are removed from the training split.
Evaluation is performed only on events from the held-out classes in the held-out test participant.

This setting is not directly supported by conventional closed-set classifiers trained only on seen classes, because their output space are independent classification probability without semantic connections between classes, which can neither explicitly include unseen labels nor interpolate from existing labels.
%Therefore, closed-set HAR baselines are marked as not applicable in this experiment.
Thus, we compare open-vocabulary and sensor-language baselines that can score or generate descriptions for class names introduced at inference time following the process in \cref{sec:evaluation_closed_set}.
As shown in Table~\ref{tab:heldout_activity_closedset}, ActNarrator achieves the strongest Accuracy and Macro-F1 on held-out activities, suggesting that its motion-token representation and sensor-language alignment transfer better to completely unseen activity classes under cross-subject evaluation.

%We first generate a natural-language activity description and then map it to the same 22-class taxonomy using a fixed deterministic LLM prompt with temperature set to zero.
%The same class list, prompt, and decoding settings are used for methods.
%We report held-out Accuracy and held-out Macro-F1 over test intervals from the held-out activity classes.

\begin{table}[h]
\centering
\scriptsize
\caption{
Held-out activity class classification under the cross-subject (XS) protocol.
%For each XS fold, all paired sensor-label or sensor-text training intervals belonging to \textit{walking}, \textit{sitting}, and \textit{jumping} are removed from the training participants, and evaluation is performed only on those held-out classes in the held-out test participant.
}
\label{tab:heldout_activity_closedset}
\begin{tabular}{lcc}
\toprule
\textbf{Method} 
& \textbf{Held-out Acc. $\uparrow$} 
& \textbf{Held-out Macro-F1 $\uparrow$} \\
\midrule
Closed-set DeepConvLSTM          
& N/A & N/A \\
Closed-set TinyHAR               
& N/A & N/A \\
Closed-set CrossHAR              
& N/A & N/A \\
\midrule
HARGPT                  
& $12.4 \pm 1.4$ & $10.7 \pm 1.2$ \\
IMU2CLIP                
& $17.3 \pm 1.3$ & $15.4 \pm 1.1$ \\
OVHAR                   
& $19.5 \pm 1.2$ & $17.4 \pm 1.1$ \\
SensorLLM               
& $21.8 \pm 1.1$ & $19.8 \pm 1.0$ \\
SensorLM                
& $23.6 \pm 1.0$ & $21.4 \pm 1.0$ \\
ActNarrator               
& $\mathbf{31.4 \pm 0.8}$ & $\mathbf{29.2 \pm 0.7}$ \\
\bottomrule
\end{tabular}
\end{table}

%Table~\ref{tab:heldout_activity_closedset} evaluates whether open-vocabulary methods can recover activity classes that were not available as paired sensor examples during training.
%This diagnostic uses standard HAR metrics for interpretability, but differs from conventional closed-set classification because held-out class names are introduced only at inference time.
%Conventional closed-set classifiers cannot produce these held-out labels without redefining and retraining their classifier heads.
%In contrast, open-vocabulary methods can score or generate activity descriptions from candidate class names.

\section{Discussion}
\label{sec:discussion}

\subsection{Key Findings}

\subsubsection{Open-ended evaluation changes what HAR systems are judged on.}
ActivityNarrated is not simply a scaled-up closed-set HAR dataset, but a different evaluation setting. 
Conventional windowed HAR models remain useful when the activity taxonomy, sensor placement, and deployment setting are well defined. 
However, when evaluation requires continuous-stream localization, variable-duration events, open-vocabulary descriptions, and cross-position generalization, fixed-window classification becomes an incomplete solution. 
This is visible in the closed-set diagnostic results. 
On ActivityNarrated, closed-set baseline classification models achieve moderate cross-subject performance and degrade substantially under cross-subject-and-position evaluation. 
This reveals weaknesses of closed-set methods that struggle to naturally express paraphrases, object-level details, goal-level descriptions, multiple valid interpretations, or activity names introduced at inference time. 
Thus, the main value of the proposed benchmark is that it enables activity understanding evaluation beyond fixed labels.

\subsubsection{Sensor representation and event grounding matter as much as language model size.}

Compared with representative closed-set HAR baselines, including DeepConvLSTM, TinyHAR, and CrossHAR, ActNarrator achieves stronger classification-oriented performance while also supporting capabilities that fixed-label models do not provide. 
Unlike windowed classifiers, ActNarrator explicitly separates sensor tokenization, variable-duration event segmentation, and sensor--language adaptation, allowing it to localize events in continuous streams and generate open-vocabulary activity descriptions rather than only assigning predefined labels.

ActNarrator also outperforms language-aligned sensor baselines, including IMU2CLIP, OVHAR, SensorLLM, SensorLM, and HARGPT, under the sensor-captioning evaluation protocol. 
These baselines were originally designed around closed-set or window-level supervision, whereas ActNarrator is trained to produce temporally grounded event captions. 
The results suggest that its gains are not due only to attaching a language model to sensor data, but to the full three-stage design: reusable motion tokenization, explicit event segmentation, and event-conditioned sensor--language alignment.

The ablations support this interpretation. 
Replacing raw features with spectral tokens improves temporal localization, caption retrieval, and closed-set diagnostic performance under both XS and XSP settings, indicating that stable motion representations are a central factor. 
The language-backbone study shows that larger SLMs improve performance up to a point, but model size alone does not explain the strongest results. 
Overall, progress in open-vocabulary wearable HAR depends on sensor representation learning and temporal grounding as much as on the choice of language backbone.

\subsubsection{Open-vocabulary outputs enable capabilities that fixed labels cannot support.}
The timeline QA and held-out activity experiments provide the clearest evidence that open-ended narration adds capabilities beyond conventional HAR. 
Closed-set timelines retain some utility for taxonomy-compatible questions, such as counting or ordering coarse movement classes, but they cannot answer semantic questions that require object-level, goal-level, or compositional information. 
In contrast, ActNarrator supports both taxonomy-compatible and semantic questions because its output preserves temporal intervals together with natural-language event descriptions.

The held-out activity experiment tests the open-vocabulary claim more directly. 
Conventional closed-set classifiers are not capable to infer unseen classes because their output spaces do not include the held-out classes. 
Open-vocabulary methods can still score or generate these activity names at inference time. 
ActNarrator achieves the strongest held-out performance, suggesting that its sensor--language alignment transfers better to activity classes without paired training examples.

\subsubsection{External transfer and missing-sensor robustness are encouraging but bounded.}

The OpenMarcie results show that the main architectural trends are not specific to ActivityNarrated. 
Spectral tokenization and event-conditioned sensor--language adaptation remain beneficial in an external industrial activity domain. 
Similarly, missing-sensor results show that training with diverse placements improves robustness when fewer sensors are available at inference time. 
For example, the model trained on all 15 positions performs substantially better with wrist-only input than a model trained only on wrist data.

\begin{figure*}[h]
\centering
\caption{
Relationship between open-vocabulary captioning quality and closed-set diagnostic performance under missing-sensor evaluation on ActivityNarrated.
Hit@5 measures caption retrieval quality, and Macro-F1 measures closed-set diagnostic classification performance.
}
\label{fig:Hit-FI}
\includegraphics[width=0.8\linewidth]{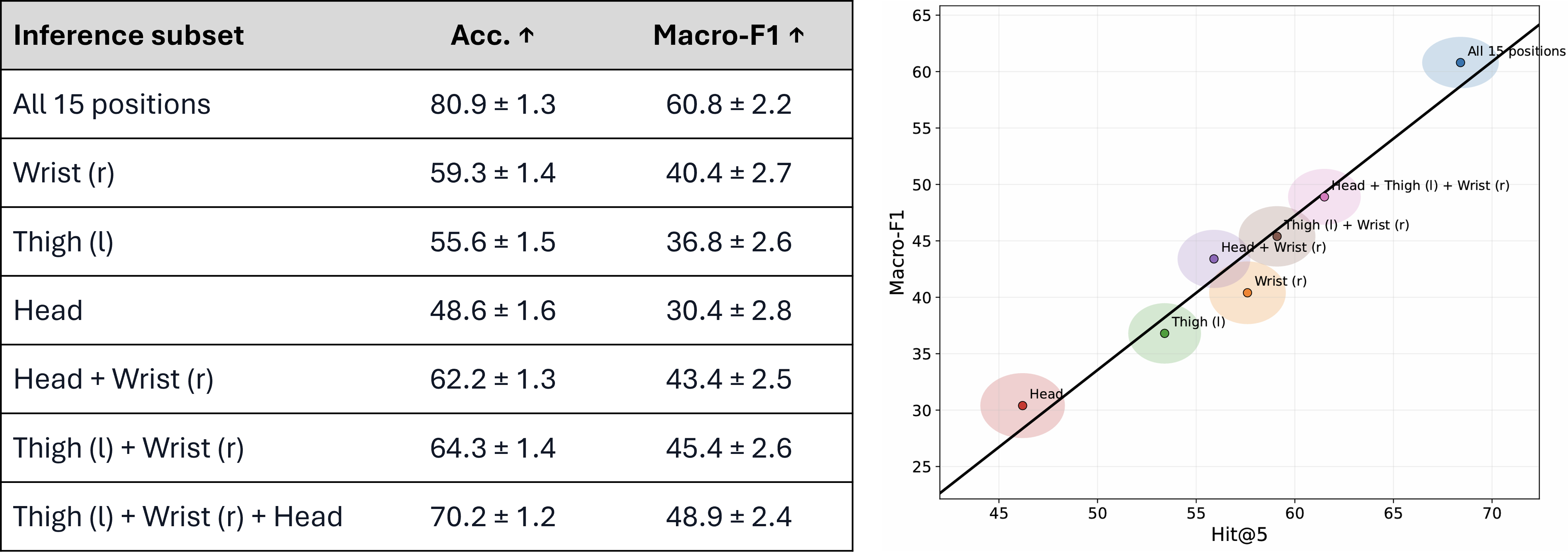}
\end{figure*}

The missing-sensor evaluation also provides a useful link between captioning quality and closed-set diagnostic performance. 
As inference sensors are reduced, captioning quality, measured by Hit@5, and closed-set classification performance, measured by Macro-F1, decrease together. 
As shown in \cref{fig:Hit-FI}, this produces a clear positive relationship between the two metrics; in this specific dataset and setting, an approximately 22-point difference in Hit@5 corresponds to about a 12-point difference in Macro-F1.

These findings support dense multi-position collection as a benchmark construction strategy, even if deployment uses a smaller sensor set. 
However, they should not be interpreted as full real-world deployment validation. 
OpenMarcie is more structured than free-living behavior, and missing-sensor evaluation does not cover all forms of device shift, long-term sensor drift, behavioral change, or environmental variation.

\subsection{Implications for Wearable Activity Understanding}

The results support a shift from treating wearable HAR only as fixed-window classification toward treating it as temporally grounded sensor--language understanding. 
The goal is not to discard closed-set recognition. 
Instead, closed-set recognition becomes one downstream view of a richer representation. 
A model that predicts intervals and captions can still be collapsed into a fixed taxonomy, but it can also support retrieval, summarization, activity-history question answering, user feedback, and human-in-the-loop review.

This matters most when the activity space is difficult to enumerate in advance. 
Assistive systems, longitudinal health monitoring, workplace activity analysis, and behavioral research often require descriptions that are more flexible than a compact fixed taxonomy. 
A fixed-label model can answer ``which known class is most likely?'' 
A narrative model can additionally answer ``what happened?'', ``when did it happen?'', ``what happened next?'', and ``does this new textual activity description match the sensor evidence?'' 
The held-out activity and timeline QA experiments demonstrate these differences in a controlled benchmark setting.

At the same time, the results show that open-vocabulary HAR is not solved by language modeling alone. 
Language provides a flexible semantic interface, but the wearable model must still learn grounded motion evidence. 
This is the central design lesson from ActNarrator: open-vocabulary activity understanding requires joint progress in data collection, sensor representation, temporal localization, and evaluation.

\subsection{Limitations and Emergent Issues}

\subsubsection{The benchmark is semi-naturalistic, not fully free-living.}
While the ultimate goal of HAR is adoption in fully unconstrained free-living environments, where the user may perform any activities anywhere without feeling observed, the challenges of data collection, annotation, ethics and privacy are still unresolved.
Nonetheless, ActivityNarrated significantly improves upon mainstream HAR data collection methods and makes a step towards the hitherto longer-term goal by relaxing scripted HAR assumptions in a structured room-scale environment with designated activity hotspots. 
This design enables dense multi-position sensing, synchronization, and high-quality annotation, but it constrains the range of contexts, objects, social settings, and long-duration routines. 
Future work should expand and stress test the paradigm on longer recordings, more diverse participants, and less controlled environments. 

\subsubsection{Narrative supervision improves semantic richness but increases annotation complexity.}
The proposed formulation gains expressiveness by using natural-language intervals, but this does not make annotation easier. 
Participant self-narration can impose cognitive load, and expert review with temporal alignment is more labor-intensive than assigning labels from a fixed taxonomy. 
Video and audio improve synchronization, annotation, and weak-label generation, but they also introduce privacy and intrusiveness concerns. 
They should be viewed as dataset-construction aids, not as requirements for wearable-only inference.
The key scalability challenge is obtaining reliable open-ended supervision without making data collection impractical. 
VLM-derived weak labels and automatic transcription may help, but the current results do not establish them as replacements for human supervision. 
Future work should explore mixed-supervision pipelines that combine expert annotation, participant descriptions, weak labels, active learning, and privacy-preserving review.

\subsubsection{Open-vocabulary evaluation still needs better semantic judgment.}
Retrieval and captioning metrics provide useful quantitative comparisons, but they remain imperfect proxies for semantic correctness. 
A retrieved description may be a valid paraphrase, a broader activity description, or a different but plausible concept, yet still be penalized if it is not included in the accepted reference set. 
Conversely, a generated caption may be linguistically similar while missing an important object, temporal relation, or goal.
The downstream QA task partially addresses this by testing whether predicted timelines support useful reasoning, and the unsupported-answer rate measures one form of hallucination. 
However, QA is still selectively controlled. 
Future evaluations should include more nuanced semantic judgment such as human-rated semantic relevance, graded paraphrase judgments, larger held-out class sets, compositional held-out phrases, and user-authored downstream questions.

\subsubsection{ActNarrator is inherently capable of adaptive activity event windows.}
Temporal granularity is an important aspect in HAR.
There are two key settings in ActNarrator's architecture defining the temporal granularity: the chunk length of the sensor tokenizer (2\,s) and the context window length to the 2nd and 3rd stage (20\,s, 10 tokens per sensor).
Since ActNarrator's event segmentation module segments activity events with a continuously differentiable scale relative to the context window, the granularity of the event's start and end is effectively infinitely small and unrestricted by the chunk granularity (e.g. an event can be 1.25\,s - 5.43\,s within the 20\,s context window).
However, the amount of sensor information that can be passed to the Q-Former is limited by the tokenizer's corresponding time window of 2\,s.
Thus, for example, in the rare occasion that within the time span of one 2\,s chunk two different short events "A" and "B" are identified, both events will be given identical sensor tokens.
Conventional HAR methods will struggle mapping the same input tokens to two distinct semantic meanings; however, the open-vocabulary narration method inherently solves this problem, as the narration ground-truth can be merged into one as "the person is doing A and/then B".

\subsubsection{Deployment requires smaller and more efficient models.}
The strongest configuration uses spectral tokenization, event segmentation, Q-Former adaptation, and frozen LLM backbones with billions of parameters. 
This is appropriate for benchmark development but not yet ideal for on-device wearable deployment. 
Practical systems will require smaller sensor-language adapters, streaming inference, distillation, model compression, and careful separation between on-device processing and optional cloud-based language reasoning.
A realistic deployment path may involve using lightweight on-device models for tokenization and event proposal generation, with larger language models used selectively for summarization, search, or user-facing explanations. 
Future work should evaluate this tradeoff explicitly for wearable deployment.

\section{Conclusion}
\label{sec:conclusion}
This work is aimed at helping the HAR community to transition from the dominant paradigm of predefined closed-set classification over fixed windows, towards an open-ended paradigm that is closer to real-world human activity understanding problems, where spatial-temporal sensor data is mapped to continuous and fluid natural language descriptions.
We introduced ActivityNarrated, a experimentation and evaluation protocol, with a collected public dataset, that instantiates and formalizes this open-ended wearable activity understanding paradigm into a concrete problem of \textbf{dense sensor signal captioning}. 
%, by representing continuous IMU streams with flexible temporal segments of natural-language descriptions. 
The dataset is collected with a semi-naturlistic setting where the participants could perform any activity on their own liberty without prescribed activities, within an environment of rich options simulating daily life scenarios.
%This enables evaluation under variable-duration behavior, cross-participant generalization, cross-position shift, missing-sensor inference, open-vocabulary semantic alignment, and downstream activity-history reasoning.

To demonstrate how this open-ended framing of HAR can be solved, we also developed a three-stage architecture  ActNarrator.% for dense sensor signal captioning with an open-vocabulary. 
The model combines discrete tokenization, query-based adaptive event segmentation, sensor to language token representation alignment, and a frozen SLM. Together, the model predicts variable-duration activity intervals and generate open-vocabulary captions. 
Unlike any HAR baseline methods, ActNarrator explicitly discovers activity events in continuous streams before captioning them, achieving accurate event segmentation with near-second boundary error.

While datasets that support this paradigm remain scarce, evaluation on both our ActivityNarrated dataset and the recent public OpenMarcie dataset shows the advantages of this formulation. 
ActNarrator improves captioning and retrieval over language-aligned baseline models, remains robust under unseen users, unseen sensor positions, missing sensors, and unseen activities, and converts wearable sensor streams into text-level activity timelines for long-horizon downstream reasoning.

The same model trained with the dense captioning task not only supports captioning, but also long-horizon timeline question answering (QA), and closed-set classification without any retraining or fine-tuning. 
In the closed-set diagnostic setting, ActNarrator improves Macro-F1 over SOTA HAR baselines by 3.8--31.6 \%. 
The long-horizon timeline QA capability at inference enables tasks like temporal ordering, temporal localization, counting of activities across the whole 1-hr recording without the need of any additional specially crafted temporal or concept hierarchical algorithms.
Overall, these results suggest a foundational style solution for wearable sensor-based HAR: train once with a generic objective, solve many downstream tasks in inference-only mode with superior performance outperforming specialized SOTA methods, turning sensor signal understanding into language-aligned reasoning, beyond backward compatability with classification, but also new ways to exploit sensor signals. %and reason over sensor data through language rather than task-specific label spaces.

\bibliographystyle{ACM-Reference-Format}
\bibliography{refs}

@article{guan2017lstmensemble,
  title={Ensembles of deep lstm learners for activity recognition using wearables},
  author={Guan, Yu and Pl{\"o}tz, Thomas},
  journal={Proceedings of the ACM on interactive, mobile, wearable and ubiquitous technologies},
  volume={1},
  number={2},
  pages={1--28},
  year={2017},
  publisher={ACM New York, NY, USA}
}

@article{abedin2021attend,
  title={Attend and discriminate: Beyond the state-of-the-art for human activity recognition using wearable sensors},
  author={Abedin, Alireza and Ehsanpour, Mahsa and Shi, Qinfeng and Rezatofighi, Hamid and Ranasinghe, Damith C},
  journal={Proceedings of the ACM on Interactive, Mobile, Wearable and Ubiquitous Technologies},
  volume={5},
  number={1},
  pages={1--22},
  year={2021},
  publisher={ACM New York, NY, USA}
}

@article{radu2018multimodal,
  title={Multimodal deep learning for activity and context recognition},
  author={Radu, Valentin and Tong, Catherine and Bhattacharya, Sourav and Lane, Nicholas D and Mascolo, Cecilia and Marina, Mahesh K and Kawsar, Fahim},
  journal={Proceedings of the ACM on interactive, mobile, wearable and ubiquitous technologies},
  volume={1},
  number={4},
  pages={1--27},
  year={2018},
  publisher={ACM New York, NY, USA}
}

@article{peng2018aroma,
  title={Aroma: A deep multi-task learning based simple and complex human activity recognition method using wearable sensors},
  author={Peng, Liangying and Chen, Ling and Ye, Zhenan and Zhang, Yi},
  journal={Proceedings of the ACM on Interactive, Mobile, Wearable and Ubiquitous Technologies},
  volume={2},
  number={2},
  pages={1--16},
  year={2018},
  publisher={ACM New York, NY, USA}
}

@inproceedings{murahari2018attentionISWC,
  title={On attention models for human activity recognition},
  author={Murahari, Vishvak S and Pl{\"o}tz, Thomas},
  booktitle={Proceedings of the 2018 ACM international symposium on wearable computers},
  pages={100--103},
  year={2018}
}

@article{tang2021selfhar,
  title={Selfhar: Improving human activity recognition through self-training with unlabeled data},
  author={Tang, Chi Ian and Perez-Pozuelo, Ignacio and Spathis, Dimitris and Brage, Soren and Wareham, Nick and Mascolo, Cecilia},
  journal={Proceedings of the ACM on interactive, mobile, wearable and ubiquitous technologies},
  volume={5},
  number={1},
  pages={1--30},
  year={2021},
  publisher={ACM New York, NY, USA}
}

@article{vaizman2018contextinthewild,
  title={Context recognition in-the-wild: Unified model for multi-modal sensors and multi-label classification},
  author={Vaizman, Yonatan and Weibel, Nadir and Lanckriet, Gert},
  journal={Proceedings of the ACM on Interactive, Mobile, Wearable and Ubiquitous Technologies},
  volume={1},
  number={4},
  pages={1--22},
  year={2018},
  publisher={ACM New York, NY, USA}
}

@article{kang2022partial,
  title={Augmented adversarial learning for human activity recognition with partial sensor sets},
  author={Kang, Hua and Huang, Qianyi and Zhang, Qian},
  journal={Proceedings of the ACM on Interactive, Mobile, Wearable and Ubiquitous Technologies},
  volume={6},
  number={3},
  pages={1--30},
  year={2022},
  publisher={ACM New York, NY, USA}
}

@article{hong2024crosshar,
  title={Crosshar: Generalizing cross-dataset human activity recognition via hierarchical self-supervised pretraining},
  author={Hong, Zhiqing and Li, Zelong and Zhong, Shuxin and Lyu, Wenjun and Wang, Haotian and Ding, Yi and He, Tian and Zhang, Desheng},
  journal={Proceedings of the ACM on Interactive, Mobile, Wearable and Ubiquitous Technologies},
  volume={8},
  number={2},
  pages={1--26},
  year={2024},
  publisher={ACM New York, NY, USA}
}

@article{chen2024sensor2text,
  title={Sensor2text: Enabling natural language interactions for daily activity tracking using wearable sensors},
  author={Chen, Wenqiang and Cheng, Jiaxuan and Wang, Leyao and Zhao, Wei and Matusik, Wojciech},
  journal={Proceedings of the ACM on Interactive, Mobile, Wearable and Ubiquitous Technologies},
  volume={8},
  number={4},
  pages={1--26},
  year={2024},
  publisher={ACM New York, NY, USA}
}

@article{wang2024ubiphysio,
  title={Ubiphysio: Support daily functioning, fitness, and rehabilitation with action understanding and feedback in natural language},
  author={Wang, Chongyang and Feng, Yuan and Zhong, Lingxiao and Zhu, Siyi and Zhang, Chi and Zheng, Siqi and Liang, Chen and Wang, Yuntao and He, Chengqi and Yu, Chun and others},
  journal={Proceedings of the ACM on Interactive, Mobile, Wearable and Ubiquitous Technologies},
  volume={8},
  number={1},
  pages={1--27},
  year={2024},
  publisher={ACM New York, NY, USA}
}

@article{leng2024imugpt2,
  title={Language-Based Cross Modality Transfer for Sensor-Based Human Activity Recognition},
  author={Leng, Zikang},
  year={2025},
  publisher={Georgia Institute of Technology}
}

@article{su2025imuzero,
  title={IMUZero: Zero-Shot Human Activity Recognition by Language-Based Cross Modality Fusion},
  author={Su, Jie and Ge, Fengtong and Wen, Zhenyu and Li, Taotao and Bai, Yang and Zhou, Yejian and Zhang, Xiaoqin},
  journal={Proceedings of the ACM on Interactive, Mobile, Wearable and Ubiquitous Technologies},
  volume={9},
  number={4},
  pages={1--28},
  year={2025},
  publisher={ACM New York, NY, USA}
}

@article{bock2024wear,
  title={Wear: An outdoor sports dataset for wearable and egocentric activity recognition},
  author={Bock, Marius and Kuehne, Hilde and Van Laerhoven, Kristof and Moeller, Michael},
  journal={Proceedings of the ACM on Interactive, Mobile, Wearable and Ubiquitous Technologies},
  volume={8},
  number={4},
  pages={1--21},
  year={2024},
  publisher={ACM New York, NY, USA}
}

@inproceedings{radford2021clip,
  title={Learning transferable visual models from natural language supervision},
  author={Radford, Alec and Kim, Jong Wook and Hallacy, Chris and Ramesh, Aditya and Goh, Gabriel and Agarwal, Sandhini and Sastry, Girish and Askell, Amanda and Mishkin, Pamela and Clark, Jack and others},
  booktitle={International conference on machine learning},
  pages={8748--8763},
  year={2021},
  organization={PmLR}
}

@article{wang2020wearingdiversity,
  title={A systematic study of unsupervised domain adaptation for robust human-activity recognition},
  author={Chang, Youngjae and Mathur, Akhil and Isopoussu, Anton and Song, Junehwa and Kawsar, Fahim},
  journal={Proceedings of the ACM on Interactive, Mobile, Wearable and Ubiquitous Technologies},
  volume={4},
  number={1},
  pages={1--30},
  year={2020},
  publisher={ACM New York, NY, USA}
}

@article{haresamudram2021cpc,
  title={Contrastive predictive coding for human activity recognition},
  author={Haresamudram, Harish and Essa, Irfan and Pl{\"o}tz, Thomas},
  journal={Proceedings of the ACM on Interactive, Mobile, Wearable and Ubiquitous Technologies},
  volume={5},
  number={2},
  pages={1--26},
  year={2021},
  publisher={ACM New York, NY, USA}
}

@article{haresamudram2022assessssl,
  title={Assessing the state of self-supervised human activity recognition using wearables},
  author={Haresamudram, Harish and Essa, Irfan and Pl{\"o}tz, Thomas},
  journal={Proceedings of the ACM on Interactive, Mobile, Wearable and Ubiquitous Technologies},
  volume={6},
  number={3},
  pages={1--47},
  year={2022},
  publisher={ACM New York, NY, USA}
}

@article{dai2024contrastsense,
  title={ContrastSense: domain-invariant contrastive learning for in-the-wild wearable sensing},
  author={Dai, Gaole and Xu, Huatao and Yoon, Hyungjun and Li, Mo and Tan, Rui and Lee, Sung-Ju},
  journal={Proceedings of the ACM on Interactive, Mobile, Wearable and Ubiquitous Technologies},
  volume={8},
  number={4},
  pages={1--32},
  year={2024},
  publisher={ACM New York, NY, USA}
}

@article{su2022bpd,
  title={Learning disentangled behaviour patterns for wearable-based human activity recognition},
  author={Su, Jie and Wen, Zhenyu and Lin, Tao and Guan, Yu},
  journal={Proceedings of the ACM on Interactive, Mobile, Wearable and Ubiquitous Technologies},
  volume={6},
  number={1},
  pages={1--19},
  year={2022},
  publisher={ACM New York, NY, USA}
}

@article{kwon2021imutube,
  title={Imutube: Automatic extraction of virtual on-body accelerometry from video for human activity recognition},
  author={Kwon, Hyeokhyen and Tong, Catherine and Haresamudram, Harish and Gao, Yan and Abowd, Gregory D and Lane, Nicholas D and Ploetz, Thomas},
  journal={Proceedings of the ACM on Interactive, Mobile, Wearable and Ubiquitous Technologies},
  volume={4},
  number={3},
  pages={1--29},
  year={2020},
  publisher={ACM New York, NY, USA}
}

@article{kwapisz2011wisdm,
  title={Activity recognition using cell phone accelerometers},
  author={Kwapisz, Jennifer R and Weiss, Gary M and Moore, Samuel A},
  journal={ACM SigKDD Explorations Newsletter},
  volume={12},
  number={2},
  pages={74--82},
  year={2011},
  publisher={ACM New York, NY, USA}
}

@inproceedings{anguita2013uci,
  title={A public domain dataset for human activity recognition using smartphones.},
  author={Anguita, Davide and Ghio, Alessandro and Oneto, Luca and Parra, Xavier and Reyes-Ortiz, Jorge Luis and others},
  booktitle={Esann},
  volume={3},
  number={1},
  pages={3--4},
  year={2013}
}

@article{reiss2012pamap2,
  title={PAMAP2 physical activity monitoring},
  author={Reiss, Attila},
  journal={UCI Machine Learning Repository},
  volume={10},
  pages={C5NW2H},
  year={2012}
}

@article{chavarriaga2013opportunity,
  title={The Opportunity challenge: A benchmark database for on-body sensor-based activity recognition},
  author={Chavarriaga, Ricardo and Sagha, Hesam and Calatroni, Alberto and Digumarti, Sundara Tejaswi and Tr{\"o}ster, Gerhard and Mill{\'a}n, Jos{\'e} del R and Roggen, Daniel},
  journal={Pattern Recognition Letters},
  volume={34},
  number={15},
  pages={2033--2042},
  year={2013},
  publisher={Elsevier}
}

@inproceedings{sztyler2016realworld,
  title={Towards real world activity recognition from wearable devices},
  author={Sztyler, Timo},
  booktitle={2017 IEEE International Conference on Pervasive Computing and Communications Workshops (PerCom Workshops)},
  pages={97--98},
  year={2017},
  organization={IEEE}
}

@inproceedings{gjoreski2018shl,
  title={Benchmarking the SHL recognition challenge with classical and deep-learning pipelines},
  author={Wang, Lin and Gjoreski, Hristijan and Ciliberto, Mathias and Mekki, Sami and Valentin, Stefan and Roggen, Daniel},
  booktitle={Proceedings of the 2018 ACM International Joint Conference and 2018 International Symposium on Pervasive and Ubiquitous Computing and Wearable Computers},
  pages={1626--1635},
  year={2018}
}

@inproceedings{vaizman2018extrasensory,
  title={Extrasensory app: Data collection in-the-wild with rich user interface to self-report behavior},
  author={Vaizman, Yonatan and Ellis, Katherine and Lanckriet, Gert and Weibel, Nadir},
  booktitle={Proceedings of the 2018 CHI conference on human factors in computing systems},
  pages={1--12},
  year={2018}
}

@article{stromback2020mmfit,
  title={Mm-fit: Multimodal deep learning for automatic exercise logging across sensing devices},
  author={Str{\"o}mb{\"a}ck, David and Huang, Sangxia and Radu, Valentin},
  journal={Proceedings of the ACM on Interactive, Mobile, Wearable and Ubiquitous Technologies},
  volume={4},
  number={4},
  pages={1--22},
  year={2020},
  publisher={ACM New York, NY, USA}
}

@inproceedings{chen2015utd,
  title={UTD-MHAD: A multimodal dataset for human action recognition utilizing a depth camera and a wearable inertial sensor},
  author={Chen, Chen and Jafari, Roozbeh and Kehtarnavaz, Nasser},
  booktitle={2015 IEEE International conference on image processing (ICIP)},
  pages={168--172},
  year={2015},
  organization={IEEE}
}

@article{minderer2023scalingovod,
  title={Scaling open-vocabulary object detection},
  author={Minderer, Matthias and Gritsenko, Alexey and Houlsby, Neil},
  journal={Advances in Neural Information Processing Systems},
  volume={36},
  pages={72983--73007},
  year={2023}
}

@inproceedings{yuan2024ovsam,
  title={Open-vocabulary sam: Segment and recognize twenty-thousand classes interactively},
  author={Yuan, Haobo and Li, Xiangtai and Zhou, Chong and Li, Yining and Chen, Kai and Loy, Chen Change},
  booktitle={European Conference on Computer Vision},
  pages={419--437},
  year={2024},
  organization={Springer}
}

@inproceedings{weng2023openvclip,
  title={Open-vclip: Transforming clip to an open-vocabulary video model via interpolated weight optimization},
  author={Weng, Zejia and Yang, Xitong and Li, Ang and Wu, Zuxuan and Jiang, Yu-Gang},
  booktitle={International conference on machine learning},
  pages={36978--36989},
  year={2023},
  organization={PMLR}
}

@article{temporalactionlocalization2024,
  title={Temporal Action Localization for Inertial-based Human Activity Recognition},
  author={Bock, Marius and Moeller, Michael and Van Laerhoven, Kristof},
  journal={Proceedings of the ACM on Interactive, Mobile, Wearable and Ubiquitous Technologies},
  volume={8},
  number={4},
  pages={1--19},
  year={2024},
  publisher={ACM New York, NY, USA}
}

@article{goat2024,
  title={Goat: A generalized cross-dataset activity recognition framework with natural language supervision},
  author={Miao, Shenghuan and Chen, Ling},
  journal={Proceedings of the ACM on Interactive, Mobile, Wearable and Ubiquitous Technologies},
  volume={8},
  number={4},
  pages={1--28},
  year={2024},
  publisher={ACM New York, NY, USA}
}

@inproceedings{moon2023imu2clip,
  title={IMU2CLIP: language-grounded motion sensor translation with multimodal contrastive learning},
  author={Moon, Seungwhan and Madotto, Andrea and Lin, Zhaojiang and Saraf, Aparajita and Bearman, Amy and Damavandi, Babak},
  booktitle={Findings of the Association for Computational Linguistics: EMNLP 2023},
  pages={13246--13253},
  year={2023}
}

@inproceedings{ray2025initial,
  title={Initial Findings on Sensor based Open Vocabulary Activity Recognition via Text Embedding Inversion},
  author={Ray, Lala Shakti Swarup and Zhou, Bo and Suh, Sungho and Lukowicz, Paul},
  booktitle={2025 IEEE International Conference on Pervasive Computing and Communications Workshops and other Affiliated Events (PerCom Workshops)},
  pages={685--688},
  year={2025},
  organization={IEEE}
}

@article{singh2020deep,
  title={Deep ConvLSTM with self-attention for human activity decoding using wearable sensors},
  author={Singh, Satya P and Sharma, Madan Kumar and Lay-Ekuakille, Aim{\'e} and Gangwar, Deepak and Gupta, Sukrit},
  journal={IEEE Sensors Journal},
  volume={21},
  number={6},
  pages={8575--8582},
  year={2020},
  publisher={IEEE}
}

@inproceedings{zhou2022tinyhar,
  title={Tinyhar: A lightweight deep learning model designed for human activity recognition},
  author={Zhou, Yexu and Zhao, Haibin and Huang, Yiran and Riedel, Till and Hefenbrock, Michael and Beigl, Michael},
  booktitle={Proceedings of the 2022 ACM International Symposium on Wearable Computers},
  pages={89--93},
  year={2022}
}

@article{wei2025one,
author = {Wei, Qingxin and Huang, Jiaming and Gao, Yi and Dong, Wei},
title = {One Model to Fit Them All: Universal IMU-based Human Activity Recognition with LLM-assisted Cross-dataset Representation},
year = {2025},
issue_date = {September 2025},
publisher = {Association for Computing Machinery},
address = {New York, NY, USA},
volume = {9},
number = {3},
url = {https://doi.org/10.1145/3749509},
doi = {10.1145/3749509},
journal = {Proc. ACM Interact. Mob. Wearable Ubiquitous Technol.},
month = sep,
articleno = {139},
numpages = {22}
}

@article{su2025semantic,
author = {Yan, Hua and Tan, Heng and Ding, Yi and Zhou, Pengfei and Namboodiri, Vinod and Yang, Yu},
title = {Large Language Model-guided Semantic Alignment for Human Activity Recognition},
year = {2025},
issue_date = {December 2025},
publisher = {Association for Computing Machinery},
address = {New York, NY, USA},
volume = {9},
number = {4},
url = {https://doi.org/10.1145/3770652},
doi = {10.1145/3770652},
journal = {Proc. ACM Interact. Mob. Wearable Ubiquitous Technol.},
month = dec,
articleno = {230},
numpages = {25}
}

@article{bello2026openmarcie,
  title={OpenMarcie: Dataset for Multimodal Action Recognition in Industrial Environments},
  author={Bello, Hymalai and Ray, Lala and Sorysz, Joanna and Suh, Sungho and Lukowicz, Paul},
  journal={arXiv preprint arXiv:2603.02390},
  year={2026}
}

@inproceedings{li2025sensorllm,
  title={Sensorllm: Aligning large language models with motion sensors for human activity recognition},
  author={Li, Zechen and Deldari, Shohreh and Chen, Linyao and Xue, Hao and Salim, Flora D},
  booktitle={Proceedings of the 2025 Conference on Empirical Methods in Natural Language Processing},
  pages={354--379},
  year={2025}
}

@article{zhang2025sensorlm,
  title={Sensorlm: Learning the language of wearable sensors},
  author={Zhang, Yuwei and Ayush, Kumar and Qiao, Siyuan and Heydari, A Ali and Narayanswamy, Girish and Xu, Maxwell A and Metwally, Ahmed A and Xu, Shawn and Garrison, Jake and Xu, Xuhai and others},
  journal={arXiv preprint arXiv:2506.09108},
  year={2025}
}

@inproceedings{ji2024hargpt,
  title={Hargpt: Are llms zero-shot human activity recognizers?},
  author={Ji, Sijie and Zheng, Xinzhe and Wu, Chenshu},
  booktitle={2024 IEEE International Workshop on Foundation Models for Cyber-Physical Systems \& Internet of Things (FMSys)},
  pages={38--43},
  year={2024},
  organization={IEEE}
}

@inproceedings{zhang2023video,
  title={Video-llama: An instruction-tuned audio-visual language model for video understanding},
  author={Zhang, Hang and Li, Xin and Bing, Lidong},
  booktitle={Proceedings of the 2023 conference on empirical methods in natural language processing: system demonstrations},
  pages={543--553},
  year={2023}
}

\appendix

\section{Metric Definitions}
\label{app:metric_definitions}

This appendix provides the formal definitions for the metrics.
Unless otherwise specified, interval-level metrics are computed after matching predicted events to ground-truth events.
Predicted intervals are filtered using a confidence threshold selected on the validation split to maximize event-level F1 at $\mathrm{tIoU}=0.5$.
For localization and captioning evaluation, each prediction is matched to the ground-truth interval with the highest temporal overlap, subject to a one-to-one matching constraint when required by the metric.
Unmatched predictions are treated as false positives for detection-style metrics, and unmatched ground-truth intervals are treated as missed events.
All reported interval-level results are averaged over evaluation folds and reported as mean $\pm$ standard deviation.
Recall, Accuracy, Macro-F1, and Hit@K are reported as percentages.
BERTScore, MRR, nDCG, reconstruction errors, and JS divergence are reported on their natural scales.

\subsection{Temporal Localization Metrics}
\label{app:metrics_localization}

Because ActNarrator operates on continuous IMU streams and predicts activity boundaries directly, evaluation must measure temporal localization quality in addition to semantic correctness.
We evaluate predicted intervals against annotated intervals using detection-style measures.

\paragraph{Temporal Intersection over Union.}
For a predicted interval $\hat{I}=[\hat{s},\hat{e}]$ and a ground-truth interval $I=[s,e]$, temporal IoU is:
\begin{equation}
    \mathrm{tIoU}(\hat{I}, I)=\frac{|\hat{I} \cap I|}{|\hat{I} \cup I|}.
\end{equation}
If two intervals do not overlap, their tIoU is zero.

\paragraph{Recall@tIoU.}
A ground-truth interval is counted as detected if at least one predicted interval overlaps it with tIoU above a threshold $\tau$:
\begin{equation}
    \mathrm{Recall}@\tau=
    \frac{1}{N}
    \sum_{i=1}^{N}
    \mathbf{1}
    \Big\{
    \exists \hat{I}\ \text{s.t.}\ \mathrm{tIoU}(\hat{I}, I_i)\ge \tau
    \Big\}.
\end{equation}
We report Recall@$0.3$, Recall@$0.5$, and Recall@$0.7$ to distinguish coarse event detection from precise boundary localization.

\paragraph{Mean Average Precision over temporal detection.}
For end-to-end temporal detection, predicted intervals are ranked by model confidence.
For each tIoU threshold $\tau \in \{0.3,0.5,0.7\}$, predictions are processed in descending confidence order and greedily matched to the unmatched ground-truth interval with the highest tIoU.
A prediction is counted as a true positive if its best unmatched ground-truth interval has tIoU at least $\tau$; otherwise, it is counted as a false positive.
Ground-truth intervals that are not matched by any prediction are false negatives.

Average Precision at threshold $\tau$, denoted $\mathrm{AP}_{\tau}$, is computed from the resulting precision--recall curve using monotonic precision-envelope interpolation:
\begin{equation}
    \mathrm{AP}_{\tau}
    =
    \sum_{n}
    (r_n-r_{n-1})\,p_{\mathrm{interp}}(r_n),
\end{equation}
where $r_n$ is the recall after the $n$-th confidence-ranked prediction and
\begin{equation}
    p_{\mathrm{interp}}(r_n)=\max_{\tilde{r}\ge r_n}p(\tilde{r}).
\end{equation}
The final temporal mean Average Precision is:
\begin{equation}
    \mathrm{mAP}
    =
    \frac{1}{|\mathcal{T}|}
    \sum_{\tau \in \mathcal{T}}
    \mathrm{AP}_{\tau},
    \qquad
    \mathcal{T}=\{0.3,0.5,0.7\}.
\end{equation}

\paragraph{Boundary error.}
For matched intervals, we report mean absolute start and end boundary error:
\begin{equation}
    \mathrm{Err}_{\mathrm{start}}
    =
    \frac{1}{N_m}
    \sum_{i=1}^{N_m}
    |s_i-\hat{s}_i|,
    \qquad
    \mathrm{Err}_{\mathrm{end}}
    =
    \frac{1}{N_m}
    \sum_{i=1}^{N_m}
    |e_i-\hat{e}_i|,
\end{equation}
where $N_m$ is the number of matched prediction--ground-truth pairs.
The single boundary-error value reported in tables, denoted $\mathrm{Bnd.\ Err.}$, is the average of start and end errors:
\begin{equation}
    \mathrm{Bnd.\ Err.}
    =
    \frac{1}{2}
    \left(
    \mathrm{Err}_{\mathrm{start}}
    +
    \mathrm{Err}_{\mathrm{end}}
    \right).
\end{equation}
Boundary error is reported in seconds.

\subsection{Captioning and Retrieval Metrics}
\label{app:metrics_caption_retrieval}

After temporal matching, we evaluate whether predicted event representations and generated captions align with the corresponding natural-language activity descriptions.
Unless otherwise specified, captioning and retrieval metrics are computed on temporally matched prediction--ground-truth pairs with $\mathrm{tIoU}\geq0.5$.
This ensures that semantic evaluation is performed on predictions that correspond to the correct temporal event, rather than rewarding semantically plausible captions attached to poorly localized intervals.

\paragraph{Candidate pool construction.}
For retrieval-style evaluation, we construct a fixed candidate text pool separately for each evaluation fold using all unique normalized activity descriptions in the test split.
Duplicate normalized descriptions are removed, so each unique candidate appears once.
The same candidate pool is used for all compared methods within a fold, and no negative subsampling is performed at test time.
We report candidate-pool size together with retrieval results, since ranking difficulty depends on the number and diversity of candidate descriptions.

\paragraph{Semantic caption similarity.}
For each temporally matched predicted interval, we compare the generated caption with the set of valid ground-truth descriptions associated with that interval.
We report BERTScore-F1 computed with \texttt{roberta-large}, using \texttt{rescale\_with\_baseline=True} and \texttt{idf=False}.
We compute BERTScore-F1 against each reference using:
\begin{equation}
    \mathrm{BERTScore}_i
    =
    \mathrm{BERTScoreF1}(\hat{y}_i,t),
\end{equation}
where $\hat{y}_i$ is the generated caption and $t$ is the valid reference descriptions for event $i$.
The final BERTScore is averaged over all matched events:
\begin{equation}
    \mathrm{BERTScore}
    =
    \frac{1}{N_m}
    \sum_{i=1}^{N_m}
    \mathrm{BERTScore}_i.
\end{equation}

\paragraph{Hit@K.}
For each matched event representation $x_i$, we compute similarity scores against every candidate description in the evaluation pool:
\begin{equation}
    s(x_i,t_j)=\cos(f(x_i),g(t_j)),
\end{equation}
where $f(\cdot)$ and $g(\cdot)$ denote the sensor-event encoder and text encoder, respectively.
Candidate descriptions are ranked in descending order by $s(x_i,t_j)$.

Let $\mathcal{T}_i$ denote the set of valid ground-truth descriptions for event $i$.
Hit@K is defined as:
\begin{equation}
    \mathrm{Hit@K}
    =
    \frac{1}{N_m}
    \sum_{i=1}^{N_m}
    \mathbf{1}
    \left\{
    \exists t \in \mathcal{T}_i
    \ \text{such that}\
    \mathrm{rank}(t \mid x_i)\le K
    \right\}.
\end{equation}
We report Hit@1 and Hit@5.

\paragraph{Mean Reciprocal Rank.}
Mean Reciprocal Rank measures how highly the first valid target description is ranked:
\begin{equation}
    \mathrm{MRR}
    =
    \frac{1}{N_m}
    \sum_{i=1}^{N_m}
    \frac{1}{\mathrm{rank}_i},
\end{equation}
where $\mathrm{rank}_i$ is the rank of the highest-ranked description in $\mathcal{T}_i$ for query $x_i$.

\paragraph{nDCG@K.}
We report normalized Discounted Cumulative Gain at rank $K$:
\begin{equation}
    \mathrm{DCG@K}
    =
    \sum_{i=1}^{K}
    \frac{2^{\mathrm{rel}_i}-1}{\log_2(i+1)},
    \qquad
    \mathrm{nDCG@K}
    =
    \frac{\mathrm{DCG@K}}{\mathrm{IDCG@K}}.
\end{equation}
We use binary relevance: $\mathrm{rel}_i=1$ if the candidate at rank $i$ is one of the valid descriptions for the event, and $\mathrm{rel}_i=0$ otherwise.
$\mathrm{IDCG@K}$ is the maximum possible DCG@K for the same number of valid descriptions.
If no valid reference appears in the candidate pool for an event, that event is excluded from retrieval evaluation.
We report nDCG@5.

\subsection{Closed-Set Diagnostic Metrics}
\label{app:metrics_classification}

Closed-set classification metrics provide comparability with conventional HAR pipelines, but they are not the primary objective of the benchmark.
For this diagnostic evaluation, activity intervals are mapped to a 23-class movement-centric taxonomy consisting of 22 canonical actions and one \textit{Other} class.

For expert-annotated intervals, we use the corresponding hard labels described in Sec.~\ref{sec:labeling}.
For generated captions, we map each caption to the same taxonomy using a fixed controlled prompt with deterministic decoding, i.e., temperature $=0$.
The same prompt, class list, and decoding settings are used for all compared methods.

\paragraph{Accuracy.}
Accuracy is the fraction of evaluated intervals assigned to the correct taxonomy class:
\begin{equation}
    \mathrm{Accuracy}
    =
    \frac{1}{N}
    \sum_{i=1}^{N}
    \mathbf{1}
    \{\hat{c}_i=c_i\},
\end{equation}
where $\hat{c}_i$ is the predicted class and $c_i$ is the ground-truth class.

\paragraph{Macro-F1.}
For each class $c$, we compute precision, recall, and F1:
\begin{equation}
    \mathrm{Precision}_c
    =
    \frac{\mathrm{TP}_c}{\mathrm{TP}_c+\mathrm{FP}_c},
    \qquad
    \mathrm{Recall}_c
    =
    \frac{\mathrm{TP}_c}{\mathrm{TP}_c+\mathrm{FN}_c},
\end{equation}
\begin{equation}
    \mathrm{F1}_c
    =
    \frac{
    2\cdot \mathrm{Precision}_c \cdot \mathrm{Recall}_c
    }{
    \mathrm{Precision}_c+\mathrm{Recall}_c
    }.
\end{equation}
Macro-F1 is the unweighted mean over the $C=23$ taxonomy classes:
\begin{equation}
    \mathrm{Macro\text{-}F1}
    =
    \frac{1}{C}
    \sum_{c=1}^{C}
    \mathrm{F1}_c.
\end{equation}
If a class has no predicted or no ground-truth instances in a fold, undefined precision or recall terms are set to zero for that fold.
Accuracy and Macro-F1 are reported as percentages.

\subsection{Representation Diagnostics}
\label{app:metrics_representation}

For token-based models such as ActNarrator, we additionally evaluate the quality of the learned discrete sensor representation.
These metrics are diagnostic rather than task-defining: they help explain downstream performance but do not directly measure activity understanding.

\paragraph{Time-domain reconstruction error.}
We report $\ell_1$ reconstruction error between the original IMU signal and the signal reconstructed from discrete motion tokens.
For an input signal $x \in \mathbb{R}^{T \times C}$ and reconstruction $\hat{x}$, time-domain error is:
\begin{equation}
    \mathrm{Time}\ \ell_1
    =
    \frac{1}{TC}
    \sum_{t=1}^{T}
    \sum_{c=1}^{C}
    |x_{t,c}-\hat{x}_{t,c}|.
\end{equation}
For multi-sensor inputs, this value is averaged over available sensors and examples.
The error is computed after the same z-score normalization used for tokenizer training, so values reflect reconstruction quality in normalized IMU space.
Lower error indicates that the tokenizer preserves more of the original motion signal.

\paragraph{Spectral reconstruction error.}
We report $\ell_1$ reconstruction error in the spectral domain to evaluate whether the tokenizer preserves periodic and transient motion structure that may be important for activity recognition.
Spectral error is computed as the average of STFT-magnitude error and CWT-magnitude error:
\begin{equation}
    \mathrm{Spectral}\ \ell_1
    =
    \frac{1}{2}
    \left(
    \frac{1}{|\mathrm{STFT}(x)|}
    \left\|
    |\mathrm{STFT}(x)|-|\mathrm{STFT}(\hat{x})|
    \right\|_1
    +
    \frac{1}{|\mathrm{CWT}(x)|}
    \left\|
    |\mathrm{CWT}(x)|-|\mathrm{CWT}(\hat{x})|
    \right\|_1
    \right).
\end{equation}
The value is averaged over frequency bins, time bins, channels, sensors, and examples.
Lower spectral error indicates better preservation of frequency-domain motion patterns.

\paragraph{Token distribution divergence.}
To measure token stability across subjects or sensor positions, we compute Jensen--Shannon divergence between token distributions.
Let $P$ and $Q$ be normalized token histograms for two conditions.
For cross-subject evaluation, $P$ and $Q$ correspond to token distributions from the training subjects and held-out subject, respectively.
For cross-position evaluation, $P$ and $Q$ correspond to token distributions from observed training positions and held-out test positions, respectively.
For missing-sensor evaluation, $P$ and $Q$ correspond to token distributions before and after restricting the available sensor subset.
Then:
\begin{equation}
    \mathrm{JS}(P \,||\, Q)
    =
    \frac{1}{2}
    \mathrm{KL}(P \,||\, M)
    +
    \frac{1}{2}
    \mathrm{KL}(Q \,||\, M),
    \qquad
    M=\frac{1}{2}(P+Q),
\end{equation}
where $\mathrm{KL}$ denotes Kullback--Leibler divergence.
Before normalization, we add $\epsilon=10^{-8}$ to each token count to avoid undefined logarithms for zero-count bins.
The reported JS value is averaged over evaluation folds.
Lower JS divergence indicates more stable token usage across the compared conditions.

\subsection{Timeline QA Metrics}
\label{app:metrics_timeline_qa}

To evaluate whether predicted activity timelines support downstream use, we report accuracy for the activity timeline question-answering task. 
A timeline consists of temporally ordered event intervals paired with either natural-language captions or closed-set class names. 
For each test session, QA pairs are generated from the human-annotated timeline using four question families: temporal ordering, time localization, counting, and semantic activity search. 
The first three families are taxonomy-compatible because they can often be answered from coarse class labels and timestamps. 
Semantic activity search requires open-vocabulary natural-language descriptions and is therefore evaluated only for timeline representations that contain captions.

All timeline sources are evaluated using the same QA procedure. 
Given a question and an input timeline, the most relevant timeline events are retrieved using sentence-embedding similarity between the question and event descriptions. 
The retrieved events are provided to the same LLM-based QA model, which is instructed to answer only from the provided timeline and to return ``unknown'' when the timeline does not contain sufficient evidence.

\paragraph{Temporal ordering QA accuracy.}
Temporal ordering questions evaluate whether the timeline preserves the relative order of activities. 
These questions ask what happened before or after a reference event, or whether one event occurred earlier than another. 
Accuracy is computed as:
\begin{equation}
    \mathrm{Temporal\ Ordering\ QA\ Acc.}
    =
    \frac{1}{N_{\mathrm{order}}}
    \sum_{i=1}^{N_{\mathrm{order}}}
    \mathbf{1}
    \{\hat{a}_i=a_i\},
\end{equation}
where $\hat{a}_i$ is the predicted answer and $a_i$ is the reference answer after deterministic normalization.

\paragraph{Time localization QA accuracy.}
Time localization questions evaluate whether the timeline supports queries about when an activity occurred. 
These questions require the model to identify the correct event interval or timestamp from the provided activity history. 
Accuracy is computed as:
\begin{equation}
    \mathrm{Time\ Localization\ QA\ Acc.}
    =
    \frac{1}{N_{\mathrm{time}}}
    \sum_{i=1}^{N_{\mathrm{time}}}
    \mathbf{1}
    \{\hat{a}_i=a_i\}.
\end{equation}

\paragraph{Counting QA accuracy.}
Counting questions evaluate whether the timeline preserves event frequency information. 
These questions ask how many times an activity or activity type occurred within the session. 
Accuracy is computed as:
\begin{equation}
    \mathrm{Counting\ QA\ Acc.}
    =
    \frac{1}{N_{\mathrm{count}}}
    \sum_{i=1}^{N_{\mathrm{count}}}
    \mathbf{1}
    \{\hat{a}_i=a_i\}.
\end{equation}

\paragraph{Semantic QA accuracy.}
Semantic activity search questions evaluate whether the timeline contains object-level, goal-level, or compositional activity descriptions. 
These questions include queries such as whether the participant interacted with food, handled an object, or performed a goal-directed action. 
Closed-set timelines are marked as N/A for this metric because their output representation contains only taxonomy-level class names and timestamps. 
Semantic QA accuracy is computed as:
\begin{equation}
    \mathrm{Semantic\ QA\ Acc.}
    =
    \frac{1}{N_{\mathrm{sem}}}
    \sum_{i=1}^{N_{\mathrm{sem}}}
    \mathbf{1}
    \{\hat{a}_i=a_i\}.
\end{equation}

\subsection{Held-Out Activity-Class Metrics}
\label{app:metrics_heldout_activity}

For held-out activity-class evaluation, we report Accuracy and Macro-F1 over test intervals belonging only to the held-out activity classes.
In this setting, selected activity classes are removed from paired sensor-label and sensor-text training examples.
The class names are introduced only at inference time through text candidates or deterministic caption-to-class mapping.

For retrieval-based open-vocabulary methods, the candidate pool consists of the 22 canonical activity class names, excluding \textit{Other}.
The class with the highest sensor--text similarity is used as the prediction.
For generative methods, the generated natural-language caption is mapped to the same class list using a fixed deterministic LLM prompt with temperature $=0$.

Accuracy is computed over held-out-class intervals:
\begin{equation}
    \mathrm{Held\text{-}out\ Acc.}
    =
    \frac{1}{N_{\mathrm{ho}}}
    \sum_{i=1}^{N_{\mathrm{ho}}}
    \mathbf{1}
    \{\hat{c}_i=c_i\},
\end{equation}
where $N_{\mathrm{ho}}$ is the number of evaluated held-out-class intervals.

Held-out Macro-F1 is computed as the unweighted mean of F1 over the held-out activity classes:
\begin{equation}
    \mathrm{Held\text{-}out\ Macro\text{-}F1}
    =
    \frac{1}{|\mathcal{C}_{\mathrm{ho}}|}
    \sum_{c\in \mathcal{C}_{\mathrm{ho}}}
    \mathrm{F1}_c,
\end{equation}
where $\mathcal{C}_{\mathrm{ho}}$ is the set of held-out activity classes.
Conventional closed-set classifiers are marked as N/A in this evaluation because their classifier heads do not include activity classes removed from the training output space.

\section{ActNarrator Implementation Details}
\label{app:actnarrator_details}

This appendix provides additional details for the ActNarrator reference architecture, including spectral tokenization, token-space augmentation, multi-sensor token encoding, dense event prediction, caption generation, training losses, and inference-time proposal merging. 
All implementation details are aligned with the experimental setup in Sec.~\ref{sec:train_config}.

\subsection{Spectral VQ-VAE Tokenizer}
\label{app:actnarrator_tokenizer}

For each sensor position, an IMU stream $x \in \mathbb{R}^{T \times C}$ with $T$ timesteps and $C$ channels is divided into non-overlapping chunks of duration $\Delta t=2.0$\,s. 
At the dataset sampling rate of approximately 30\,Hz, each chunk contains approximately 60 samples. 
We use $C=9$ channels per sensor: 3-axis acceleration, 3-axis linear acceleration, and 3-axis gyroscope. 
This yields $M$ chunks $\{x^{(m)}\}_{m=1}^{M}$. 
Each chunk is encoded and mapped to a discrete codebook entry, producing a temporally ordered token sequence with one token every 2\,s.

All IMU channels are z-score normalized using statistics computed on the training split. 
The same normalization statistics are applied to validation and test splits.

\paragraph{Multi-view spectral representation.}
Each IMU chunk is represented using three complementary views: the raw time-domain signal, short-time Fourier transform (STFT) magnitude features, and continuous wavelet transform (CWT) magnitude features. 
The time-domain view preserves local motion amplitude and direction, while the spectral views capture periodic and transient motion structure. 
This is useful for distinguishing activities such as walking, jumping, reaching, lifting, and object manipulation, where both local waveform shape and frequency content are informative.

Let an IMU segment be represented as a multivariate time series:
\[
\mathbf{x}(t) = [x_1(t), x_2(t), \dots, x_C(t)]^\top,
\qquad
t=0,\dots,T-1,
\]
where $C$ is the number of IMU channels and $T$ is the number of time steps in the segment.

For each channel $c$, the STFT is computed as:
\begin{equation}
\mathrm{STFT}_c(\tau,\omega)
=
\sum_{t=0}^{L-1}
x_c(t+\tau) w(t) e^{-j\omega t},
\end{equation}
where $\tau$ is the local window index, $L$ is the window length, $w(t)$ is the windowing function, and $\omega$ is frequency. 
We use the magnitude $|\mathrm{STFT}_c(\tau,\omega)|$ as the frequency-domain input.

The CWT is computed as:
\begin{equation}
\mathrm{CWT}_c(a,b)
=
\frac{1}{\sqrt{|a|}}
\sum_{t=0}^{T-1}
x_c(t)
\psi^*
\left(
\frac{t-b}{a}
\right),
\end{equation}
where $a>0$ is the scale parameter, $b$ is the temporal translation, $\psi$ is the mother wavelet, and $^*$ denotes complex conjugation. 
The CWT provides a multi-scale time-frequency representation that is useful for non-stationary and transient IMU patterns.

After applying STFT or CWT, the resulting spectral matrices from all channels are stacked:
\begin{equation}
\mathbf{X}_{\mathrm{spec}}
=
[\mathrm{Spec}_1,\dots,\mathrm{Spec}_C]
\in \mathbb{R}^{C \times F \times T'},
\end{equation}
where $F$ is the number of frequency bins or wavelet scales, and $T'$ is the number of time frames after windowing. 
STFT captures localized frequency components, while CWT provides a flexible multi-scale view. 
The two spectral views are therefore complementary for IMU tokenization.

\paragraph{Multi-view encoders.}
The time-domain signal is processed using a 1D convolutional encoder. 
The STFT and CWT representations are processed using separate 2D convolutional encoders. 
The resulting time, STFT, and CWT embeddings are concatenated and projected to a latent vector $r_m \in \mathbb{R}^{d}$ for each chunk.

\paragraph{Vector quantization.}
Given a codebook $E=\{e_k\}_{k=1}^{K}$ with $K=128$ entries, vector quantization assigns each chunk to its nearest code:
\begin{equation}
    z_m = \arg\min_{k \in \{1,\dots,K\}} \|r_m - e_k\|_2,
    \qquad
    q_m = e_{z_m}.
\end{equation}
The discrete token $z_m$ is passed to the downstream event-captioning model, while the quantized embedding $q_m$ is used for reconstruction during tokenizer training.

\paragraph{Tokenizer reconstruction objective.}
The decoder reconstructs the time-domain IMU signal and spectral targets from the quantized embeddings. 
The tokenizer is trained with:
\begin{equation}
\mathcal{L}_{\mathrm{tok}} =
\mathcal{L}_{\mathrm{time}}
+ \lambda_{\mathrm{stft}} \mathcal{L}_{\mathrm{stft}}
+ \lambda_{\mathrm{cwt}} \mathcal{L}_{\mathrm{cwt}}
+ \mathcal{L}_{\mathrm{codebook}}
+ \mathcal{L}_{\mathrm{commit}}.
\end{equation}

The reconstruction terms are:
\begin{align}
\mathcal{L}_{\mathrm{time}} 
&= 
\frac{1}{M}\sum_{m=1}^{M}
\left\|x^{(m)} - \hat{x}^{(m)}\right\|_1, \\
\mathcal{L}_{\mathrm{stft}} 
&= 
\frac{1}{M}\sum_{m=1}^{M}
\left\|
|\mathrm{STFT}(x^{(m)})| - |\mathrm{STFT}(\hat{x}^{(m)})|
\right\|_1, \\
\mathcal{L}_{\mathrm{cwt}} 
&= 
\frac{1}{M}\sum_{m=1}^{M}
\left\|
|\mathrm{CWT}(x^{(m)})| - |\mathrm{CWT}(\hat{x}^{(m)})|
\right\|_1.
\end{align}

The vector-quantization and commitment terms are:
\begin{align}
\mathcal{L}_{\mathrm{codebook}} 
&=
\frac{1}{M}\sum_{m=1}^{M}
\left\|
\mathrm{sg}[r_m] - q_m
\right\|_2^2, \\
\mathcal{L}_{\mathrm{commit}} 
&=
\frac{\beta}{M}\sum_{m=1}^{M}
\left\|
r_m - \mathrm{sg}[q_m]
\right\|_2^2,
\end{align}
where $\mathrm{sg}[\cdot]$ denotes the stop-gradient operator. 
The spectral losses encourage the tokenizer to retain periodic and transient motion patterns that may be attenuated by time-domain reconstruction alone.

\paragraph{Tokenizer hyperparameters.}
The tokenizer uses the following reference settings:
\begin{itemize}
    \item chunk duration: $\Delta t=2.0$\,s;
    \item token stride: $2.0$\,s;
    \item number of IMU channels: $C=9$;
    \item codebook size: $K=128$;
    \item optimizer: Adam;
    \item learning rate: $1\times10^{-3}$;
    \item batch size: 64;
    \item weight decay: $1\times10^{-4}$;
    \item maximum epochs: 300;
    \item early stopping patience: 10.
\end{itemize}

\subsection{Token-Based Augmentations}
\label{app:actnarrator_token_aug}

Discrete IMU tokens represent short reusable motion primitives, or micro-actions, that compose into higher-level activities. 
To improve robustness to natural execution variability, sensor noise, and partial observability, we apply stochastic augmentations directly in token space during dense event-captioning training. 
These augmentations are applied only during training and are disabled during validation and evaluation.

We use three token-space augmentations. 
First, repeated-token removal removes short runs of consecutive repeated tokens, simulating temporal contraction of sustained micro-actions and reducing over-reliance on exact duration. 
Second, token insertion adds tokens sampled either from nearby token neighborhoods or from the global token distribution, simulating temporal expansion, spurious observations, or sensor noise. 
Third, segment swapping exchanges short contiguous subsequences within a local temporal neighborhood, modeling small local reorderings of micro-actions without changing the global activity semantics.

These augmentations preserve the discrete-token interface while increasing compositional variability. 
They regularize the temporal encoder and event-query module, improve robustness under cross-subject and cross-position shifts, and help the model tolerate missing or noisy sensor observations.

\begin{figure*}[t]
    \centering
    \includegraphics[width=0.8\linewidth]{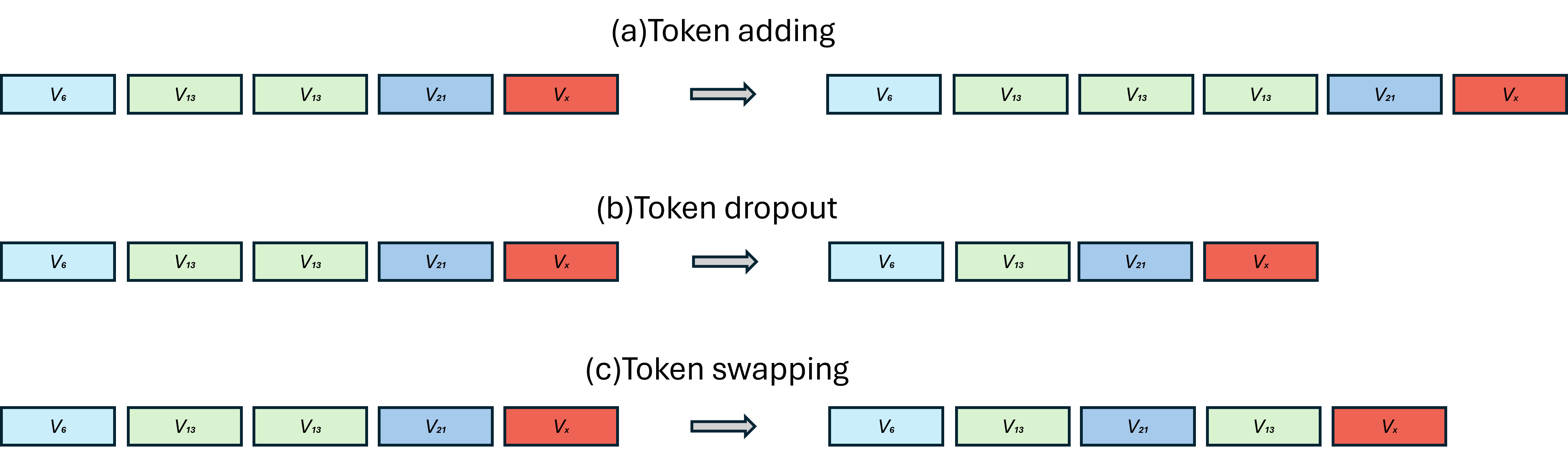}
    \caption{
    Token-based augmentations for IMU sequences.
    Discrete sensor-token sequences are augmented using repeated-token removal, token insertion, and local segment swapping.
    These operations increase compositional variability and encourage robust sensor--language alignment.
    }
    \label{fig:token_augmentations}
\end{figure*}

\subsection{Event Queries and Temporal Boundary Prediction}
\label{app:actnarrator_event_queries}

To detect multiple events within a context window, the model uses $Q$ learnable event queries $\{a_q\}_{q=1}^{Q}$. 
Each query represents a candidate activity event within $T_{\mathrm{ctx}}=20$\,s. 
The queries are processed by an event query decoder, where they interact with one another through self-attention and attend to the encoded multi-position IMU token timeline $H$ through cross-attention. 
The decoder produces one event representation $r_q$ for each query.

The event representation $r_q$ is passed to temporal proposal heads. 
The confidence head predicts whether the query corresponds to a valid event:
\begin{equation}
    \hat{p}_q \in [0,1].
\end{equation}

The boundary head predicts a normalized event center and duration:
\begin{equation}
    \hat{c}_q \in [0,1],
    \qquad
    \hat{d}_q \in [0,1].
\end{equation}

The normalized start and end times are obtained from the center-duration parameterization:
\begin{equation}
    \hat{s}_q = \max\left(0, \hat{c}_q - \frac{\hat{d}_q}{2}\right),
    \qquad
    \hat{e}_q = \min\left(1, \hat{c}_q + \frac{\hat{d}_q}{2}\right).
\end{equation}

These normalized boundaries are converted to seconds by scaling with the context-window duration $T_{\mathrm{ctx}}=20$\,s and adding the context-window offset. 
Although the input token sequence is produced at 2\,s resolution, the boundary head regresses continuous start and end times, so the predicted boundaries are not restricted to multiples of 2\,s.

Each valid proposal defines an event interval:
\[
    T_q^{\mathrm{event}}=[s_q,e_q],
\]
and the corresponding event-specific token subset:
\[
    Z_q^{\mathrm{event}} \subseteq \{Z^{(p)}_{1:M_p}\}_{p \in \mathcal{P}_{\mathrm{obs}}},
\]
where $Z_q^{\mathrm{event}}$ contains the encoded sensor tokens that fall within the predicted temporal span.

In addition, an event identity head maps each event representation $r_q$ to an identity embedding:
\begin{equation}
    v_q = f_{\mathrm{id}}(r_q).
\end{equation}
This embedding captures the event-level motion pattern and temporal context of the proposal. 
During inference, it is used to compare proposals from overlapping context windows, so that temporally overlapping proposals with similar identity embeddings can be treated as duplicate detections and merged during temporal non-maximum suppression.

\subsection{Q-Former Adaptation and Caption Generation}
\label{app:actnarrator_captioning}

After dense event localization, each valid event proposal $k$ is represented by a temporal interval $T_k=[s_k,e_k]$ and an event-specific token subset $Z_k$. 
The subset $Z_k$ contains the encoded IMU tokens from the observed sensor positions whose timestamps fall inside the predicted interval. 
This stage generates the semantic description for each localized event; it does not modify the predicted event boundaries.

The event-specific token sequence $Z_k$ is passed to a lightweight Q-Former, which maps the selected sensor-token sequence into the embedding space of a frozen decoder-only language model. 
In our formulation, the Q-Former preserves the event-token structure and produces a semantic token sequence $Z'_k$ with the same token count as the selected event tokens. 
Therefore, if event $k$ contains $n$ temporal tokens from each of $p$ observed sensor positions, then $Z'_k$ contains $n \times p$ semantic tokens.

Each input token in $Z_k$ retains the temporal and sensor-position information introduced during multi-sensor token encoding. 
The resulting semantic token sequence $Z'_k$ is combined with a short textual header that specifies the predicted interval and the available sensor positions:
\begin{quote}\footnotesize
\texttt{Describe the activity between $s_k$ and $e_k$ seconds using IMU evidence from \{pos\_1, pos\_2, ...\}.}
\end{quote}

This header makes the temporal span and sensor availability explicit, supporting evaluation under different sensor-position subsets. 
The frozen language model generates one open-vocabulary activity caption conditioned on both the semantic sensor tokens $Z'_k$ and the textual header. 
The language model parameters remain frozen; gradients from the captioning objective update only the trainable sensor encoder, event modules, Q-Former, and projection layers.

For each matched event query, caption generation is trained using a next-token prediction loss ($\mathcal{L}_{\mathrm{cap}}$):
\begin{equation}
\mathcal{L}_{\mathrm{cap}} =
-\sum_{k \in \mathcal{M}}
\sum_{t=1}^{L_k}
\log p_\theta
\left(
y_t^{(k)}
\mid
y_{<t}^{(k)}, \mathrm{header}_k, Z'_k
\right),
\end{equation}
where $\mathcal{M}$ is the set of event proposals matched to ground-truth intervals, $y^{(k)}$ is the target caption for event $k$, and $L_k$ is the caption length. 
The notation $\theta$ denotes the trainable non-LLM parameters; the decoder-only language model weights remain frozen. 
Unmatched no-event queries do not receive caption loss.

In the main experiments, we use Qwen~2.5~7B as the frozen decoder-only language model because it provides the strongest validation performance among the evaluated backbones. 
We additionally evaluate Qwen~2.5~1.5B, Qwen~2.5~3B, Gemma~2B, and LLaMA~8B variants in the language-backbone ablation.
\subsection{Hungarian Matching and Training Objective}
\label{app:actnarrator_training}

Ground-truth annotations are represented as intervals with start time, end time, and one or more natural-language descriptions. 
During training, predicted event queries are assigned to ground-truth intervals using one-to-one Hungarian matching. 
This matching is necessary because the $K$ event queries form an unordered set of candidate event proposals.

For predicted query $k$ and ground-truth event $j$, the matching cost is:
\begin{equation}
\mathcal{C}_{kj} =
\lambda_{\mathrm{conf}}
\left[-\log(\hat{p}_k)\right]
+ \lambda_{\ell_1}
\left(
|\hat{s}_k - s_j| + |\hat{e}_k - e_j|
\right)
+ \lambda_{\mathrm{tiou}}
\left(1 - \mathrm{tIoU}(\hat{I}_k, I_j)\right),
\end{equation}
where $\hat{I}_k=[\hat{s}_k,\hat{e}_k]$ is the predicted interval and $I_j=[s_j,e_j]$ is the ground-truth interval.

Matched queries are optimized with event confidence and boundary localization losses. 
The event confidence loss encourages matched queries to predict valid activity events, while the boundary localization loss penalizes errors in the predicted temporal center and duration, or equivalently the converted start and end times. 
Unmatched queries are trained with a no-event loss so that unused queries do not produce false proposals.

The event identity head is trained with an identity consistency loss. 
Proposals matched to the same ground-truth event are encouraged to have similar identity embeddings, while proposals matched to different ground-truth events are encouraged to remain separated in the embedding space. 
This encourages the identity embedding to represent event-level uniqueness, making it useful for duplicate-aware merging across overlapping context windows.

For matched event queries, the corresponding event-specific token subset $Z_k$ is passed to the captioning module, and caption generation is optimized with a next-token prediction loss. 
Unmatched no-event queries do not receive caption loss.

The full training objective is:
\begin{equation}
\mathcal{L}_{\mathrm{dense}} =
\lambda_{\mathrm{evt}} \mathcal{L}_{\mathrm{evt}}
+ \lambda_{\mathrm{loc}} \mathcal{L}_{\mathrm{loc}}
+ \lambda_{\mathrm{id}} \mathcal{L}_{\mathrm{id}}
+ \lambda_{\mathrm{cap}} \mathcal{L}_{\mathrm{cap}}.
\end{equation}
Here, $\mathcal{L}_{\mathrm{evt}}$ is the event/no-event classification loss, $\mathcal{L}_{\mathrm{loc}}$ combines boundary regression and temporal-overlap losses for matched queries, $\mathcal{L}_{\mathrm{id}}$ supervises event identity embeddings, and $\mathcal{L}_{\mathrm{cap}}$ is applied only to matched event-caption pairs.

\subsection{Inference and Temporal Non-Maximum Suppression}
\label{app:actnarrator_inference}

At inference time, the model is applied over overlapping context windows of duration $T_{\mathrm{ctx}}=20$\,s across the full IMU stream. 
Each window produces up to $K$ event proposals:
\[
    (\hat{s}_k, \hat{e}_k, \hat{p}_k, v_k, \hat{y}_k),
\]
where $\hat{s}_k$ and $\hat{e}_k$ are predicted start and end times, $\hat{p}_k$ is event confidence, $v_k$ is the event identity embedding, and $\hat{y}_k$ is the generated activity caption.

Low-confidence proposals are removed using a confidence threshold selected on the validation split. 
Remaining proposals from overlapping context windows are merged using temporal non-maximum suppression. 
Proposals are sorted by confidence, and a lower-confidence proposal is suppressed if it has high temporal overlap with a higher-confidence proposal and a similar event identity embedding. 
This duplicate-aware merging treats temporally overlapping and semantically similar proposals as detections of the same underlying activity event.

For each retained event proposal, the event-specific token subset $Z_k$ is passed to the Q-Former and frozen language model to generate an activity caption. 
Caption decoding uses deterministic decoding with temperature 0. 
Generated captions are lowercased and stripped of leading/trailing whitespace before evaluation, but no semantic post-processing is applied.

The retained proposals are sorted by start time to form the final dense activity narration. 
Because temporal NMS suppresses highly overlapping proposals, this inference procedure focuses on a dominant temporally ordered event sequence. 
Future extensions could retain overlapping proposals to model simultaneous or nested activity narratives.

\section{Additional Evaluation Results}
\label{app:additional_results}

This appendix reports additional ablations and expanded results that complement Sec.~\ref{sec:evaluation}. 
The main paper focuses on the primary ActivityNarrated results, while the appendix provides tokenizer ablations, LLM-backbone scaling, detailed OpenMarcie results, closed-set missing-sensor diagnostics, and timeline-QA construction details.

\subsection{Tokenizer Ablations}
\label{app:tokenizer_ablations}

Table~\ref{tab:dict_ablation_app} reports reconstruction error and token divergence for different codebook sizes.
Increasing dictionary size improves reconstruction, but very large dictionaries reduce token reuse across subjects and positions.
We use $K=128$ in subsequent experiments because it provides the best balance between reconstruction fidelity and cross-position token stability.

\begin{table}[t]
\caption{
Ablation over dictionary size $K$ on ActivityNarrated and OpenMarcie.
ActivityNarrated results are reported under XS and XSP splits.
OpenMarcie is reported separately as an external XS setting.
}
\centering
\scriptsize
\setlength{\tabcolsep}{5pt}
\begin{tabular}{c ccc}
\toprule
\textbf{$K$}
& \textbf{Time $\ell_1$} $\downarrow$
& \textbf{Spectral $\ell_1$} $\downarrow$
& \textbf{JS} $\downarrow$ \\
\midrule
\multicolumn{4}{l}{\textit{ActivityNarrated -- XS}} \\
\midrule
32  & 0.118 $\pm$ 0.006 & 0.164 $\pm$ 0.009 & 0.41 $\pm$ 0.03 \\
64  & 0.097 $\pm$ 0.005 & 0.138 $\pm$ 0.007 & 0.31 $\pm$ 0.02 \\
\textbf{128} & \textbf{0.082 $\pm$ 0.004} & \textbf{0.116 $\pm$ 0.006} & \textbf{0.22 $\pm$ 0.02} \\
256 & 0.079 $\pm$ 0.004 & 0.113 $\pm$ 0.006 & 0.29 $\pm$ 0.03 \\
512 & 0.076 $\pm$ 0.003 & 0.111 $\pm$ 0.005 & 0.38 $\pm$ 0.04 \\
\midrule
\multicolumn{4}{l}{\textit{ActivityNarrated -- XSP}} \\
\midrule
32  & 0.132 $\pm$ 0.008 & 0.181 $\pm$ 0.011 & 0.47 $\pm$ 0.04 \\
64  & 0.111 $\pm$ 0.007 & 0.155 $\pm$ 0.009 & 0.36 $\pm$ 0.03 \\
\textbf{128} & \textbf{0.095 $\pm$ 0.006} & \textbf{0.131 $\pm$ 0.008} & \textbf{0.27 $\pm$ 0.02} \\
256 & 0.092 $\pm$ 0.006 & 0.129 $\pm$ 0.007 & 0.35 $\pm$ 0.03 \\
512 & 0.090 $\pm$ 0.005 & 0.127 $\pm$ 0.007 & 0.44 $\pm$ 0.04 \\
\midrule
\multicolumn{4}{l}{\textit{OpenMarcie -- XS}} \\
\midrule
32  & 0.146 $\pm$ 0.008 & 0.201 $\pm$ 0.011 & 0.52 $\pm$ 0.04 \\
64  & 0.123 $\pm$ 0.007 & 0.173 $\pm$ 0.010 & 0.41 $\pm$ 0.03 \\
\textbf{128} & \textbf{0.106 $\pm$ 0.006} & \textbf{0.148 $\pm$ 0.008} & \textbf{0.31 $\pm$ 0.03} \\
256 & 0.103 $\pm$ 0.006 & 0.145 $\pm$ 0.008 & 0.39 $\pm$ 0.03 \\
512 & 0.101 $\pm$ 0.005 & 0.143 $\pm$ 0.007 & 0.48 $\pm$ 0.04 \\
\bottomrule
\end{tabular}
\label{tab:dict_ablation_app}
\end{table}

Table~\ref{tab:length_ablation_app} evaluates discretization windows from 1\,s to 5\,s.
A 2\,s window consistently provides the best tradeoff.
Shorter windows fragment coherent motion patterns, while longer windows blur multiple motion phases into a single token.
We therefore use a 2\,s discretization window in downstream experiments.

\begin{table}[t]
\caption{
Ablation over discretization window length $\Delta t$ on ActivityNarrated and OpenMarcie.
Intermediate durations (2\,s) yield the best tradeoff between reconstruction quality and compositional token structure.
}
\centering
\scriptsize
\setlength{\tabcolsep}{5pt}
\begin{tabular}{c ccc}
\toprule
\textbf{$\Delta t$ (s)}
& \textbf{Time $\ell_1$} $\downarrow$
& \textbf{Spectral $\ell_1$} $\downarrow$
& \textbf{JS} $\downarrow$ \\
\midrule
\multicolumn{4}{l}{\textit{ActivityNarrated -- XS}} \\
\midrule
1.0  & 0.091 $\pm$ 0.005 & 0.128 $\pm$ 0.007 & 0.29 $\pm$ 0.03 \\
\textbf{2.0} & \textbf{0.082 $\pm$ 0.004} & \textbf{0.116 $\pm$ 0.006} & \textbf{0.22 $\pm$ 0.02} \\
3.0  & 0.087 $\pm$ 0.005 & 0.123 $\pm$ 0.007 & 0.31 $\pm$ 0.03 \\
5.0  & 0.104 $\pm$ 0.006 & 0.149 $\pm$ 0.009 & 0.42 $\pm$ 0.04 \\
\midrule
\multicolumn{4}{l}{\textit{ActivityNarrated -- XSP}} \\
\midrule
1.0  & 0.105 $\pm$ 0.007 & 0.145 $\pm$ 0.009 & 0.34 $\pm$ 0.03 \\
\textbf{2.0} & \textbf{0.095 $\pm$ 0.006} & \textbf{0.131 $\pm$ 0.008} & \textbf{0.27 $\pm$ 0.02} \\
3.0  & 0.101 $\pm$ 0.007 & 0.139 $\pm$ 0.009 & 0.37 $\pm$ 0.03 \\
5.0  & 0.121 $\pm$ 0.009 & 0.168 $\pm$ 0.011 & 0.48 $\pm$ 0.05 \\
\midrule
\multicolumn{4}{l}{\textit{OpenMarcie -- XS}} \\
\midrule
1.0  & 0.116 $\pm$ 0.007 & 0.162 $\pm$ 0.009 & 0.38 $\pm$ 0.03 \\
\textbf{2.0} & \textbf{0.106 $\pm$ 0.006} & \textbf{0.148 $\pm$ 0.008} & \textbf{0.31 $\pm$ 0.03} \\
3.0  & 0.112 $\pm$ 0.007 & 0.157 $\pm$ 0.009 & 0.40 $\pm$ 0.03 \\
5.0  & 0.134 $\pm$ 0.008 & 0.188 $\pm$ 0.011 & 0.52 $\pm$ 0.04 \\
\bottomrule
\end{tabular}
\label{tab:length_ablation_app}
\end{table}

\begin{figure*}[t]
    \centering
    \includegraphics[width=0.9\linewidth]{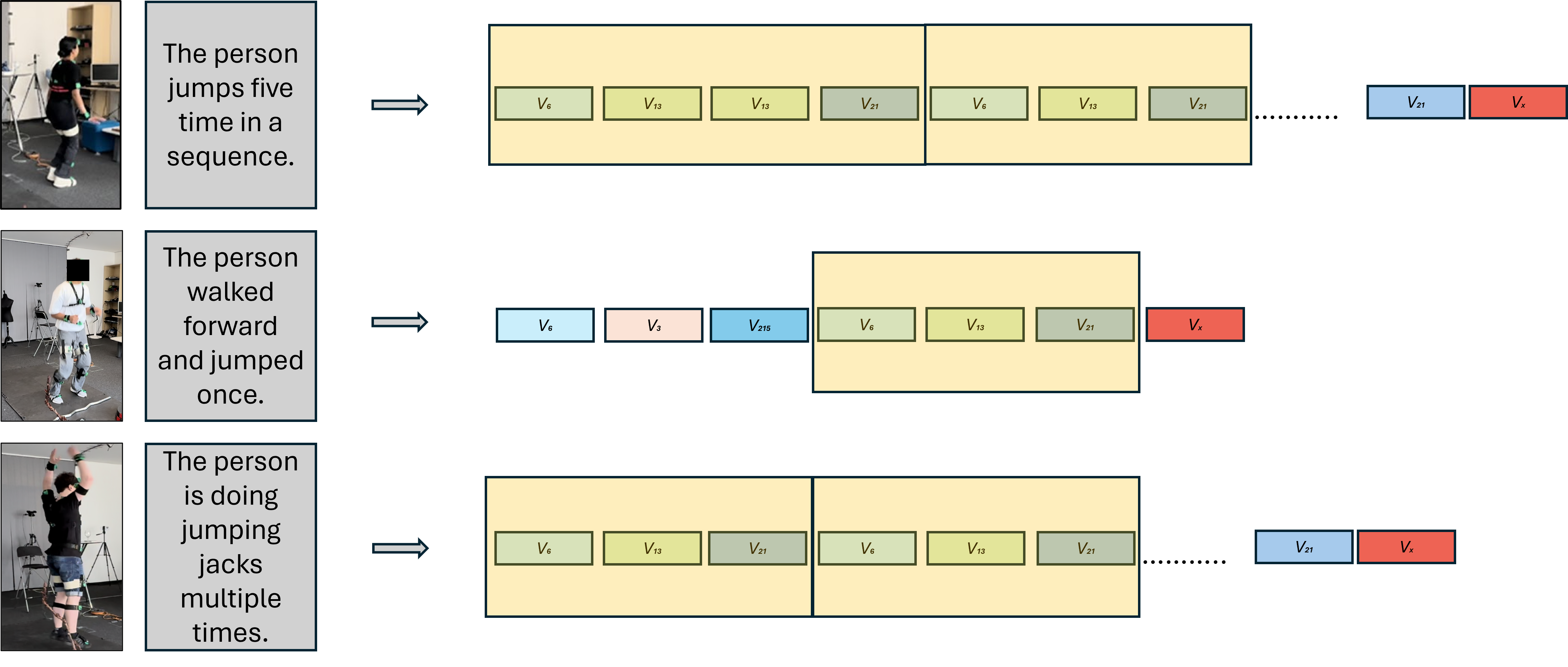}
    \caption{
    Micro-activity decomposition via IMU tokenization.
    Similar activities yield recurring token patterns that repeat over time, illustrating how extended behaviors can be represented as sequences of shared local motion patterns.
    }
    \label{fig:micro_activity_tokens_app}
\end{figure*}

Figure~\ref{fig:micro_activity_tokens_app} visualizes token sequences over extended recordings.
Recurring token patterns appear across repeated or similar motions, suggesting that the tokenizer decomposes longer behaviors into reusable local motion patterns.
This qualitative structure is consistent with the lower JS divergence observed for the Spectral VQ-VAE under XS and XSP evaluation.

\subsection{LLM Backbone Scaling}
\label{app:llm_backbone_scaling}

Table~\ref{tab:ov_llm_activitynarrated_openmarcie_app} reports the effect of frozen LLM backbone choice.
Performance generally improves from smaller models to Qwen~2.5~7B.
LLaMA~8B does not outperform Qwen~2.5~7B in this setup, suggesting that larger language capacity alone is not sufficient; the sensor--language adapter and event representation remain central bottlenecks.

\begin{table*}[t]
\caption{
Impact of frozen language model backbone architecture and parameter scale on dense open-vocabulary event captioning performance.
}
\centering
\scriptsize
\setlength{\tabcolsep}{5pt}
\begin{tabular}{l ccccc}
\toprule
\textbf{Backbone}
& \textbf{BERTScore $\uparrow$}
& \textbf{Hit@1 $\uparrow$}
& \textbf{Hit@5 $\uparrow$}
& \textbf{MRR $\uparrow$}
& \textbf{nDCG@5 $\uparrow$} \\
\midrule
\multicolumn{6}{l}{\textit{ActivityNarrated -- XS}} \\
\midrule
Qwen 2.5 1.5B
& 0.688 $\pm$ 0.010 & 31.4 $\pm$ 1.0 & 61.2 $\pm$ 1.8 & 0.49 $\pm$ 0.03 & 0.59 $\pm$ 0.03 \\
Gemma 2B
& 0.680 $\pm$ 0.010 & 30.6 $\pm$ 0.9 & 59.8 $\pm$ 1.7 & 0.48 $\pm$ 0.03 & 0.58 $\pm$ 0.03 \\
Qwen 2.5 3B
& 0.697 $\pm$ 0.009 & 32.4 $\pm$ 1.0 & 62.7 $\pm$ 1.8 & 0.50 $\pm$ 0.03 & 0.60 $\pm$ 0.03 \\
\textbf{Qwen 2.5 7B}
& \textbf{0.734 $\pm$ 0.008} & \textbf{36.2 $\pm$ 1.1} & \textbf{68.4 $\pm$ 1.9} & \textbf{0.55 $\pm$ 0.03} & \textbf{0.65 $\pm$ 0.04} \\
LLaMA 8B
& 0.704 $\pm$ 0.009 & 33.0 $\pm$ 1.0 & 64.1 $\pm$ 1.8 & 0.51 $\pm$ 0.03 & 0.61 $\pm$ 0.03 \\
\midrule
\multicolumn{6}{l}{\textit{ActivityNarrated -- XSP}} \\
\midrule
Qwen 2.5 1.5B
& 0.672 $\pm$ 0.011 & 28.5 $\pm$ 0.9 & 56.7 $\pm$ 1.7 & 0.46 $\pm$ 0.03 & 0.55 $\pm$ 0.03 \\
Gemma 2B
& 0.663 $\pm$ 0.011 & 27.7 $\pm$ 0.8 & 55.0 $\pm$ 1.6 & 0.45 $\pm$ 0.03 & 0.54 $\pm$ 0.03 \\
Qwen 2.5 3B
& 0.682 $\pm$ 0.010 & 29.6 $\pm$ 0.9 & 58.2 $\pm$ 1.7 & 0.47 $\pm$ 0.03 & 0.56 $\pm$ 0.03 \\
\textbf{Qwen 2.5 7B}
& \textbf{0.712 $\pm$ 0.009} & \textbf{32.5 $\pm$ 1.0} & \textbf{63.1 $\pm$ 1.9} & \textbf{0.52 $\pm$ 0.03} & \textbf{0.61 $\pm$ 0.04} \\
LLaMA 8B
& 0.688 $\pm$ 0.010 & 30.1 $\pm$ 0.9 & 59.3 $\pm$ 1.7 & 0.48 $\pm$ 0.03 & 0.57 $\pm$ 0.03 \\
\midrule
\multicolumn{6}{l}{\textit{OpenMarcie -- XS}} \\
\midrule
Qwen 2.5 1.5B
& 0.655 $\pm$ 0.014 & 35.6 $\pm$ 1.3 & 64.1 $\pm$ 2.3 & 0.49 $\pm$ 0.04 & 0.60 $\pm$ 0.04 \\
Gemma 2B
& 0.649 $\pm$ 0.014 & 34.5 $\pm$ 1.3 & 62.3 $\pm$ 2.2 & 0.48 $\pm$ 0.04 & 0.59 $\pm$ 0.04 \\
Qwen 2.5 3B
& 0.662 $\pm$ 0.013 & 36.8 $\pm$ 1.3 & 65.7 $\pm$ 2.3 & 0.50 $\pm$ 0.04 & 0.61 $\pm$ 0.04 \\
\textbf{Qwen 2.5 7B}
& \textbf{0.701 $\pm$ 0.012} & \textbf{40.4 $\pm$ 1.4} & \textbf{72.0 $\pm$ 2.4} & \textbf{0.56 $\pm$ 0.04} & \textbf{0.68 $\pm$ 0.04} \\
LLaMA 8B
& 0.668 $\pm$ 0.013 & 37.1 $\pm$ 1.3 & 66.2 $\pm$ 2.3 & 0.51 $\pm$ 0.04 & 0.62 $\pm$ 0.04 \\
\bottomrule
\end{tabular}
\label{tab:ov_llm_activitynarrated_openmarcie_app}
\end{table*}

\subsection{Detailed OpenMarcie Results}
\label{app:openmarcie_detailed}

OpenMarcie~\cite{bello2026openmarcie} is used as an external-domain benchmark.
It differs from ActivityNarrated in environment, task structure, and activity distribution.
Table~\ref{tab:openmarcie_tokenizer_app} compares IMU tokenization strategies on OpenMarcie under the XS setting, evaluating time-domain reconstruction error, spectral reconstruction error, and token-distribution stability.

\begin{table}[t]
\caption{
OpenMarcie tokenizer comparison under the XS setting.
}
\centering
\scriptsize
\setlength{\tabcolsep}{5pt}
\begin{tabular}{l ccc}
\toprule
\textbf{Tokenizer}
& \textbf{Time $\ell_1$} $\downarrow$
& \textbf{Spectral $\ell_1$} $\downarrow$
& \textbf{JS} $\downarrow$ \\
\midrule
Raw VQ-VAE & 0.409 $\pm$ 0.024 & 0.548 $\pm$ 0.031 & 0.64 $\pm$ 0.04 \\
STFT VQ-VAE & 0.184 $\pm$ 0.013 & 0.263 $\pm$ 0.017 & 0.56 $\pm$ 0.05 \\
Wavelet VQ-VAE & 0.194 $\pm$ 0.014 & 0.247 $\pm$ 0.016 & 0.49 $\pm$ 0.05 \\
\textbf{Spectral VQ-VAE} & \textbf{0.106 $\pm$ 0.006} & \textbf{0.148 $\pm$ 0.008} & \textbf{0.31 $\pm$ 0.03} \\
\bottomrule
\end{tabular}
\label{tab:openmarcie_tokenizer_app}
\end{table}

Table~\ref{tab:openmarcie_temporal_app} reports dense temporal localization results on OpenMarcie, measuring event recall, temporal mAP, and boundary error.

\begin{table}[t]
\caption{
OpenMarcie temporal localization under the XS setting.
}
\centering
\scriptsize
\setlength{\tabcolsep}{5pt}
\begin{tabular}{l ccccc}
\toprule
\textbf{Method}
& \textbf{Rec@0.3} $\uparrow$
& \textbf{Rec@0.5} $\uparrow$
& \textbf{Rec@0.7} $\uparrow$
& \textbf{mAP} $\uparrow$
& \textbf{Bnd. Err.} $\downarrow$ \\
\midrule
Raw + mean pooling & 28.4 $\pm$ 3.1 & 18.6 $\pm$ 2.7 & 8.9 $\pm$ 1.9 & 12.7 $\pm$ 2.1 & 3.42 $\pm$ 0.31 \\
Raw + attention & 31.2 $\pm$ 3.0 & 21.4 $\pm$ 2.6 & 10.8 $\pm$ 1.9 & 15.1 $\pm$ 2.1 & 3.21 $\pm$ 0.29 \\
Raw + Q-Former & 36.8 $\pm$ 2.8 & 26.7 $\pm$ 2.5 & 14.9 $\pm$ 1.8 & 20.6 $\pm$ 2.0 & 2.84 $\pm$ 0.26 \\
Spectral + mean pooling & 39.5 $\pm$ 2.7 & 29.3 $\pm$ 2.4 & 16.5 $\pm$ 1.8 & 22.8 $\pm$ 1.9 & 2.66 $\pm$ 0.24 \\
Spectral + attention & 43.6 $\pm$ 2.6 & 33.4 $\pm$ 2.3 & 20.1 $\pm$ 1.7 & 27.0 $\pm$ 1.9 & 2.38 $\pm$ 0.22 \\
Spectral + Q-Former & 49.2 $\pm$ 2.4 & 38.8 $\pm$ 2.2 & 24.7 $\pm$ 1.7 & 32.9 $\pm$ 1.8 & 2.08 $\pm$ 0.19 \\
\textbf{ActNarrator} & \textbf{53.5 $\pm$ 2.3} & \textbf{42.6 $\pm$ 2.1} & \textbf{27.8 $\pm$ 1.6} & \textbf{36.9 $\pm$ 1.7} & \textbf{1.91 $\pm$ 0.18} \\
\bottomrule
\end{tabular}
\label{tab:openmarcie_temporal_app}
\end{table}

Table~\ref{tab:openmarcie_semantic_app} evaluates dense open-vocabulary captioning and sensor--language retrieval on OpenMarcie using an 83-description candidate pool.

\begin{table*}[t]
\caption{
OpenMarcie dense open-vocabulary event captioning under the XS setting.
The candidate pool contains 83 normalized descriptions.
}
\centering
\scriptsize
\setlength{\tabcolsep}{5pt}
\begin{tabular}{l ccccc}
\toprule
\textbf{Method}
& \textbf{BERTScore} $\uparrow$
& \textbf{Hit@1} $\uparrow$
& \textbf{Hit@5} $\uparrow$
& \textbf{MRR} $\uparrow$
& \textbf{nDCG@5} $\uparrow$ \\
\midrule
IMU2CLIP & 0.505 $\pm$ 0.018 & 18.6 $\pm$ 1.2 & 43.8 $\pm$ 2.0 & 0.29 $\pm$ 0.03 & 0.41 $\pm$ 0.03 \\
OVHAR & 0.548 $\pm$ 0.016 & 23.9 $\pm$ 1.1 & 51.5 $\pm$ 2.1 & 0.36 $\pm$ 0.03 & 0.48 $\pm$ 0.03 \\
SensorLLM & 0.579 $\pm$ 0.015 & 27.2 $\pm$ 1.2 & 56.1 $\pm$ 2.1 & 0.40 $\pm$ 0.03 & 0.52 $\pm$ 0.03 \\
SensorLM & 0.604 $\pm$ 0.014 & 30.1 $\pm$ 1.2 & 60.4 $\pm$ 2.2 & 0.44 $\pm$ 0.03 & 0.56 $\pm$ 0.03 \\
HARGPT & 0.529 $\pm$ 0.017 & 20.5 $\pm$ 1.1 & 47.2 $\pm$ 2.0 & 0.32 $\pm$ 0.03 & 0.44 $\pm$ 0.03 \\
Raw + Q-Former & 0.596 $\pm$ 0.014 & 29.3 $\pm$ 1.2 & 59.0 $\pm$ 2.2 & 0.43 $\pm$ 0.03 & 0.55 $\pm$ 0.03 \\
Spectral + Q-Former & 0.672 $\pm$ 0.012 & 36.9 $\pm$ 1.3 & 68.1 $\pm$ 2.4 & 0.52 $\pm$ 0.04 & 0.64 $\pm$ 0.03 \\
\textbf{ActNarrator} & \textbf{0.701 $\pm$ 0.012} & \textbf{40.4 $\pm$ 1.4} & \textbf{72.0 $\pm$ 2.4} & \textbf{0.56 $\pm$ 0.04} & \textbf{0.68 $\pm$ 0.04} \\
\bottomrule
\end{tabular}
\label{tab:openmarcie_semantic_app}
\end{table*}

\subsection{Timeline QA Construction and Evaluation}
\label{app:timeline_qa_eval_details}

This appendix describes the downstream activity-history question-answering task used in Sec.~\ref{sec:timeline_qa}.

\paragraph{Timeline representation.}
A timeline is represented as an ordered list of events:
\[
    \mathcal{L}
    =
    \{(s_i,e_i,d_i)\}_{i=1}^{M},
\]
where $s_i$ and $e_i$ are the start and end timestamps of event $i$, and $d_i$ is either a natural-language caption or a closed-set taxonomy label.
Open-vocabulary timelines contain natural-language descriptions.
Closed-set timelines contain only taxonomy-level class names and timestamps.

\paragraph{Question categories.}
We evaluate two broad categories of questions.
Taxonomy-compatible questions can be answered from closed-set class labels and timestamps.
They include counting, ordering, timing, and duration questions over canonical movement classes.
Open-vocabulary semantic questions require natural-language event descriptions.
They include object-interaction questions, goal-level activity questions, compositional activity questions, and questions about semantically specific actions that collapse into the same closed-set class.

\paragraph{Question generation.}
Questions are generated using deterministic templates over held-out human annotations.
For taxonomy-compatible questions, templates are instantiated from the 23-class taxonomy and event timestamps.
For semantic questions, templates are instantiated from normalized natural-language descriptions and event intervals.
Questions whose answers are ambiguous under the reference annotations are excluded.
The same question set is used for all compared timeline sources within a fold.

\paragraph{Answering procedure.}
For each method, predicted event timelines are provided to the same downstream QA model.
The answerer receives only the predicted timeline, not the original sensor stream or ground-truth annotations.
This isolates whether the predicted timeline representation contains sufficient temporal and semantic information to support downstream reasoning.

Given a question and an input timeline, we retrieve the most relevant timeline events using sentence-embedding similarity between the question and event descriptions.

\paragraph{Example QA prompt.}
The following example illustrates the prompt format:

\begin{quote}
\small
You are answering questions using only the activity timeline below.

Rules:
\begin{itemize}
    \item Use only the events shown in the timeline.
    \item Do not use outside knowledge.
    \item If the timeline does not contain enough information to answer, respond with \texttt{unknown}.
    \item For counting questions, return only the number unless a short phrase is needed.
    \item For timing questions, return the timestamp or time interval in seconds.
    \item For yes/no questions, answer \texttt{yes}, \texttt{no}, or \texttt{unknown}.
    \item For open-ended questions, answer with a concise phrase grounded in the timeline.
\end{itemize}

Timeline:
\begin{verbatim}
[0.0s--4.2s] standing near the table
[4.3s--8.7s] reaching toward a cup
[8.8s--12.1s] picking up the cup
[12.2s--16.4s] drinking from the cup
[16.5s--19.6s] placing the cup back on the table
\end{verbatim}

Question:
\begin{verbatim}
Why did the participant went to the table?
\end{verbatim}

Answer:
\begin{verbatim}
To drink water
\end{verbatim}
\end{quote}

\subsection{Open-Vocabulary Inference Examples}

\begin{figure*}[t]
    \centering
    \includegraphics[width=\linewidth]{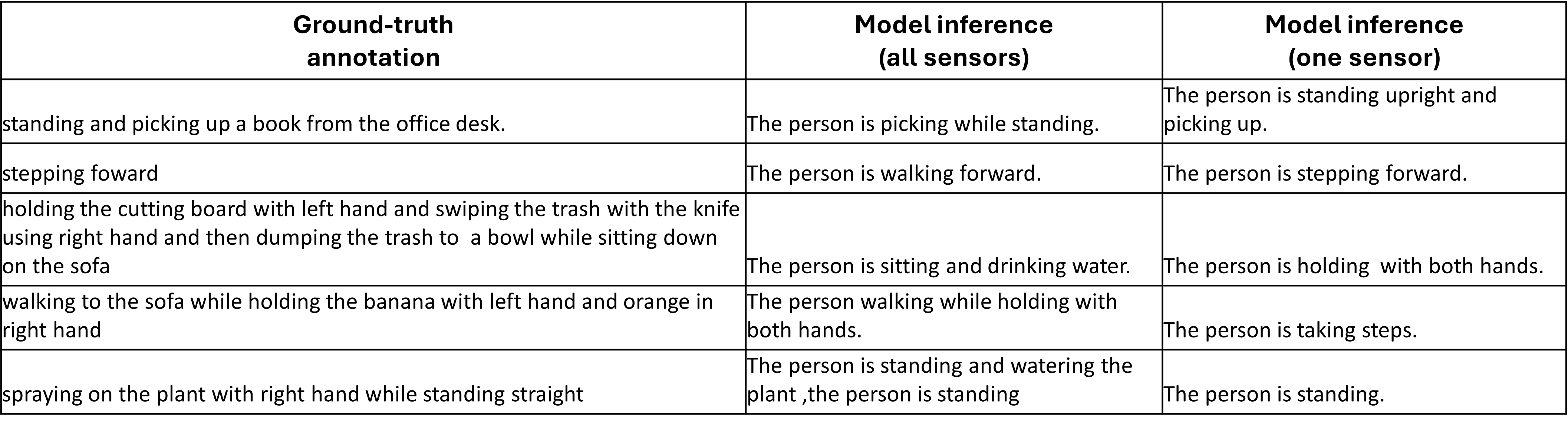}
    \caption{
    Example of ground-truth annotations vs. ActNarrator predictions for all 15 vs 1 sensor input.
    }
    \label{fig:qual}
\end{figure*}

Figure~\ref{fig:qual} illustrates examples of ground-truth annotations compared to predictions made by \textbf{ActNarrator} using either all 15 sensors or just a single sensor. These examples highlight the open-vocabulary capabilities of our framework and its robustness to missing or sparse sensor inputs.

\end{document}